\documentclass[a4paper,10pt,twocolumn]{article}

\usepackage[dvipsnames,table,xcdraw]{xcolor}

\usepackage{hegroup}

\usepackage{graphicx}
\usepackage{multirow}
\usepackage{tabularx}
\usepackage{array}
\usepackage{makecell}
\usepackage{xspace}
\usepackage{listings}
\usepackage{enumitem}
\usepackage{bbm}
\usepackage[most]{tcolorbox}
\usepackage{lipsum}
\usepackage{fontawesome5}
\usepackage{wrapfig}
\usepackage{wasysym}

\usepackage{cleveref}

\newif\ifshowchanges
\showchangestrue

\newcommand{\mypara}[1]{\noindent{\bf {#1}.}\xspace}
\newcommand{\bench}{\textit{VisAssistDaily}\xspace}
\newcommand{\trainingdata}{\textit{SafeVid}\xspace}

\begin{document}
%----------------------------------
\title{``I Can See Forever!'': Evaluating Real-time VideoLLMs for Assisting Individuals with Visual Impairments}
\author{
Ziyi Zhang\thanks{Equal contribution.}
  \ \ \ 
Zhen Sun\footnotemark[1]  \ \ \ 
Zongmin Zhang  \ \ \ 
Zifan Peng     \ \ \
Yuemeng Zhao     \ \ \ \\
Zichun Wang     \ \ \
Zeren Luo     \ \ \
Ruiting Zuo     \ \ \
Xinlei He\thanks{Corresponding author (\href{mailto:xinleihe@hkust-gz.edu.cn}{xinleihe@hkust-gz.edu.cn}).} 
\ \ \
\\
\\
\textit{The Hong Kong University of Science and Technology (Guangzhou)} \ \ \ 
\\
\\
}
\date{}   
\maketitle
%----------------------------------
\begin{abstract}
%----------------------------------

The visually impaired population faces significant challenges in daily activities. 
While prior works employ vision language models for assistance, most focus on static content and cannot address real-time perception needs in complex environments. 
Recent VideoLLMs enable real-time vision and speech interaction, offering promising potential for assistive tasks. 
In this work, we conduct the first study evaluating their effectiveness in supporting daily life for visually impaired individuals.
We first conducted a user survey with visually impaired participants to design the benchmark \bench for daily life evaluation. 
Using \bench, we evaluate popular VideoLLMs and find GPT-4o achieves the highest task success rate. 
We further conduct a user study to reveal concerns about hazard perception. To address this, we propose \trainingdata, an environment-awareness dataset, and fine-tune VITA-1.5, improving risk recognition accuracy from 25.00\% to 76.00\%.
We hope this work provides valuable insights and inspiration for future research in this field.

%----------------------------------
\end{abstract}
%----------------------------------

\begin{figure*}[t]
    \centering
    \includegraphics[width=\textwidth]{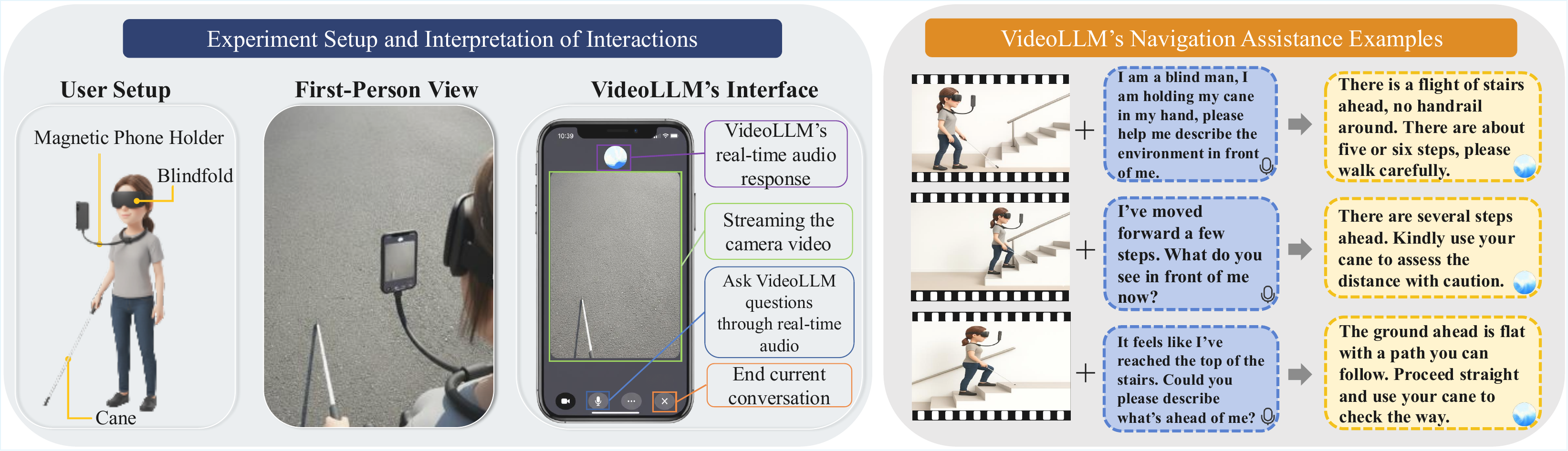}
    \caption{Overview of the experiment setup and interaction flow with examples of VideoLLM.}
    \label{fig:overview}
\end{figure*}

%----------------------------------
\section{Introduction}
%----------------------------------

With global population growth and aging, the number of people with visual impairments continues to rise.
According to the findings by the GBD 2019 Blindness and Visually Impaired Collaborators~\cite{Bourne2021Trends}, by 2050, there will be approximately 61 million blind individuals and 474 million individuals with moderate to severe visual impairments worldwide, posing an increasing challenge to public health and social infrastructure.
For society, visual impairment and blindness can lead to economic burdens~\cite{Eckert2015Cost}, reduced access to education and employment opportunities~\cite{Frick2003BlindnessCost}, and a heightened risk of mortality~\cite{mccarty2001vision}. 
For individuals, it significantly affects quality of life~\cite{taipale2019low}, may trigger mental health issues~\cite{heesterbeek2017incidence}, and increases the risks like cognitive decline~\cite{whitson2007combined} and falls~\cite{tricco2017comparisons}.

Prior works have developed numerous devices and software aimed at providing support for individuals with severe visual impairments~\cite{DBLP:conf/chi/WeiRGNZOJN25}.
Based on their functionality, these tools can be categorized into four classes: navigation aids~\cite{wewalk,nfb_sunu_band,DBLP:conf/chi/KuribayashiUWMA25,DBLP:conf/chi/Cai0GSCWHZH24,DBLP:conf/chi/KuribayashiI0VR23}, image recognition applications~\cite{OrCam,BeMyEyes,DBLP:conf/chi/PenuelaCBA24}, daily life assistive products~\cite{HumanWare}, and software-based tools~\cite{freedomscientific_jaws}. 
While these technologies offer valuable support to visually impaired individuals, they share a common limitation: the inability to truly understand the semantics of the environment or contextualized tasks.
The possession of powerful language understanding capabilities of Large Language Models (LLMs) and Vision Language Models (VLMs) offers a new solution to this problem.
With the emergence of LLMs and VLMs, many researches have begun exploring their potential in assisting visually impaired individuals~\cite{DBLP:conf/chi/Zhang24a,DBLP:journals/corr/abs-2402-01735,DBLP:conf/icaiic/KuzdeuovMNKV24,DBLP:conf/assets/AdninD24,DBLP:conf/chi/Xie0ZBL025}.
However, most of them focus on understanding and describing static image scenes without fully addressing a more crucial capability in real-world scenarios, i.e., continuous perception.
In practice, visually impaired individuals receive information through dynamic and ever-changing inputs, such as ongoing video streams.
Therefore, it is essential for models to handle sequential visual inputs and deliver assistance that is both temporally consistent and practically useful.
This gap limits the practical usability of assistive systems in complex and ever-changing environments.

The emergence of Video-based Large Language Models (VideoLLMs)~\cite{DBLP:journals/corr/abs-2312-17432} presents a promising opportunity to address these gaps.
Trained on large-scale video-text pairs, VideoLLMs have demonstrated powerful video understanding capabilities. 
Currently, VideoLLMs can be broadly categorized into offline and real-time modes. 
The latter, exemplified by models such as GPT-4o~\cite{Hurst2024GPT4oSC} and VITA-1.5~\cite{DBLP:journals/corr/abs-2501-01957} exemplify this progress. 
These models enable real-time vision and speech interaction, laying the groundwork for their application in real-world assistive tasks for visually impaired individuals.\footnote{In this paper, all mentions of VideoLLMs refer to real-time VideoLLMs.}
To evaluate whether state-of-the-art (SOTA) real-time VideoLLMs can play a substantive role in the daily life scenarios of visually impaired individuals, we conduct the first systematic empirical study.

%----------------------------------
\subsection{Our Work}
%----------------------------------

To the best of our knowledge, we present the first systematic evaluation of real-time VideoLLMs' effectiveness in assisting visually impaired individuals (as shown in~\Cref{fig:overview}).
Firstly, based on established evaluation standards~\cite{perkins_om_standards,cbra_orientation_training_2024} and our user survey with visually impaired individuals (including a questionnaire and short semi-structured interviews), we categorize the evaluation dimensions into three groups:
(1) Basic Skills (including orientation skills, guided walking, independent walking, and cane techniques);
(2) Home Life Tasks (such as housework, leisure, and recreation);
and (3) Social Life Tasks (including road walking, transportation, and reaching destinations).
For each task, we design corresponding test scenarios and task completion goals, resulting in the creation of a dataset named \bench (Benchmark for Daily Assistance to the Visually Impaired).
We select three VideoLLMs capable of real-time vision and speech interaction for bilingual (English and Chinese) evaluation: GPT-4o~\cite{Hurst2024GPT4oSC}, VITA-1.5~\cite{DBLP:journals/corr/abs-2501-01957}, and Zhipu~\cite{zhipuai2025}.
We consider four metrics in our evaluation: Task Success Rate, Prompt Cost, Response Latency, and Language Consistency.
Experiments show that GPT-4o performs the best across all metrics, demonstrating a high task success rate and strong adaptability to both English and Chinese inputs. 
In contrast, VITA-1.5 records the lowest task success rate and performs poorly in language consistency, with instances of responding in Chinese to English prompts.

In addition, we recruit visually impaired volunteers of different ages, genders, and levels of vision loss to participate in our user study.
Guided by the three evaluation dimensions defined earlier, the volunteers complete both indoor and outdoor tasks and provide structured feedback on the performance of VideoLLMs. 
The results indicate that users are generally satisfied with the response speed, ease of use, and affordability, but express concerns about accuracy in complex environments, particularly in stair recognition. 
Based on previous evaluations, we identify key challenges and future directions in this domain.
To address the ``Proactive Perception'' challenge, we construct a novel environment-awareness dataset, \trainingdata, which incorporates diverse real-world hazard scenarios such as approaching obstacles and contact with dangerous objects. 
We then fine-tune VITA-1.5 on \trainingdata for the environmental risk recognition task, improving accuracy from 25.00\% to 76.00\% and demonstrating the effectiveness of our approach in real-world applications.
In conclusion, we make the following contributions:

\begin{itemize}
    \item We conduct the first systematic evaluation of the practical effectiveness of VideoLLMs in assisting visually impaired individuals with daily life. To support this, we carry out a user survey with visually impaired individuals and use the insights to design a benchmark, \bench, which measures VideoLLMs' performances in multilingual scenarios in terms of task completion and response quality.
    \item We conduct a user study with visually impaired participants of varying backgrounds and degrees of vision loss. Through their real-world interactions with VideoLLMs, we collect authentic feedback that reveals both the strengths and limitations of these models in practical assistive scenarios.
    \item Through in-depth analysis of the experimental process and user feedback, we identify key challenges that VideoLLMs currently face in this domain. To address the most important challenge, lack of proactive prompting in VideoLLMs, we propose a dataset \trainingdata, and then fine-tuned VITA-1.5 on the environmental risk recognition task, improving accuracy from 25.00\% to 76.00\%.
\end{itemize}

\begin{figure*}[t]
    \centering
    \includegraphics[width=0.80\textwidth]{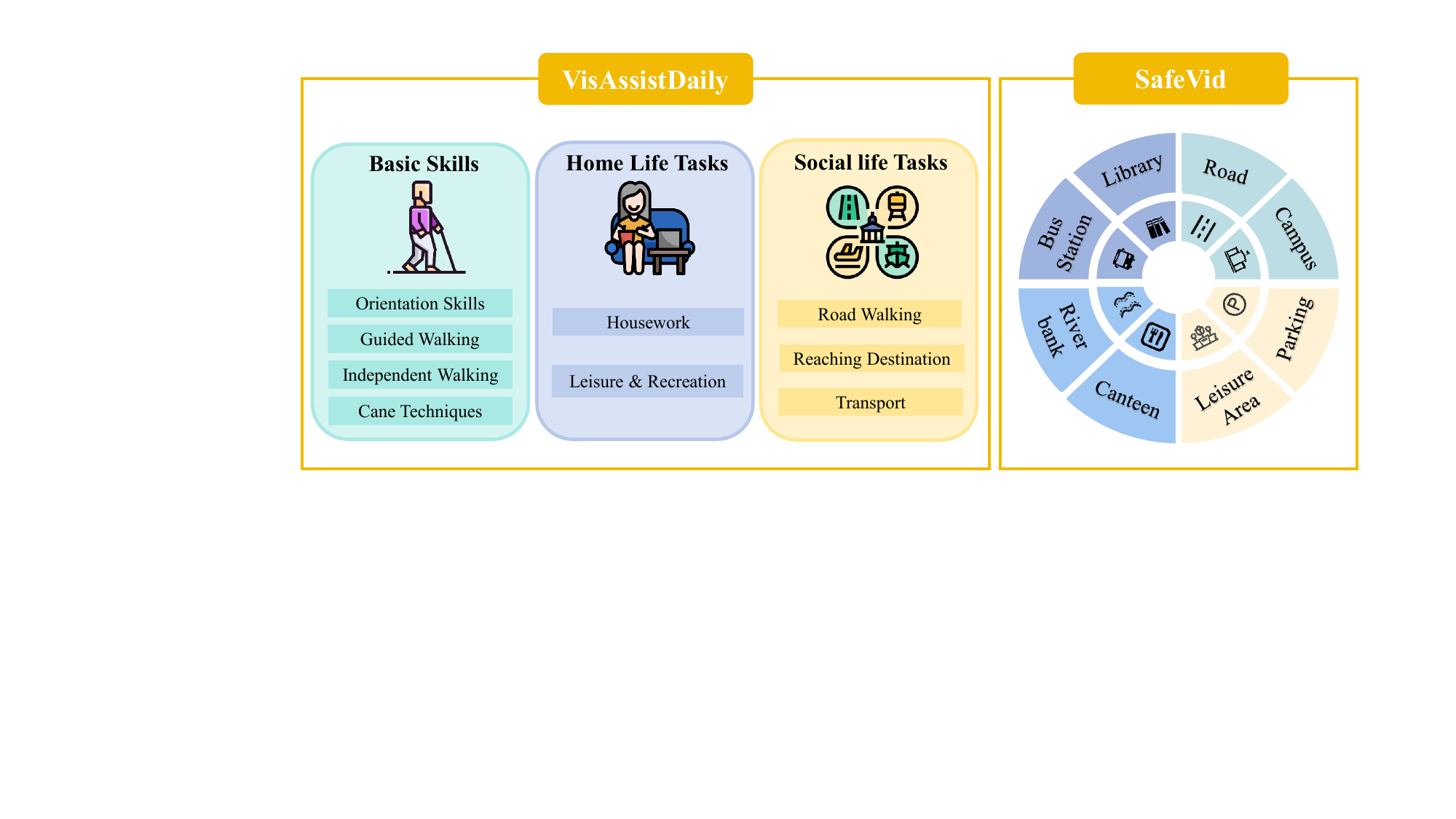}
    \caption{Overview of \bench and \trainingdata.}
    \label{fig:dataset}
\end{figure*}

%----------------------------------
\section{Preliminary and Related Work}
%----------------------------------

%----------------------------------
\subsection{Video-based Large Language Models}
%----------------------------------

With the remarkable text comprehension capabilities of LLMs, Video-based Large Language Models (VideoLLMs) have also demonstrated strong video understanding abilities after being trained on multimodal video data~\cite{DBLP:journals/corr/abs-2312-17432}. 
A common training paradigm for VideoLLMs involves using a visual encoder and a projection layer to map video inputs into the text latent space of LLMs, followed by leveraging a pre-trained LLM for text generation~\cite{DBLP:conf/eccv/WengHHCZ24}.
Video understanding is generally categorized into two types based on the mode of interaction: offline video understanding~\cite{Wu2023VisualCT} and online video understanding~\cite{DBLP:conf/cvpr/ChenLWLSGLGMS24}. 
Offline video understanding has become relatively mature within VideoLLMs and has shown strong performance~\cite{DBLP:conf/eccv/HuangLRYMYK24, DBLP:conf/emnlp/ZhangLB23, DBLP:journals/corr/abs-2305-06355}, whereas online video understanding is typically used in scenarios such as autonomous driving~\cite{Cui2023ReceiveRA} and human-computer interaction~\cite{Bi2023MISARAM}.
The need to process continuous video streams in real-time poses significantly greater challenges~\cite{DBLP:journals/corr/abs-2312-17432}.

In this paper, we focus on three commonly used models for online video inference: GPT-4o~\cite{Hurst2024GPT4oSC}, Zhipu Qingyan~\cite{zhipuai2025}, and VITA-1.5~\cite{DBLP:journals/corr/abs-2501-01957}. 
GPT-4o~\cite{Hurst2024GPT4oSC} achieves native multimodal integration, enabling unified processing of text, images, audio, and video.
It supports video input and real-time conversation, and is available on both mobile apps and web platforms, offering broad prospects for the development of multimodal interaction systems. 
Zhipu Qingyan~\cite{zhipuai2025}, abbreviated as Zhipu, is an AI application developed based on the GLM series models~\cite{Zeng2024ChatGLMAF}.
It features functions such as content generation and information summarization and also provides users with real-time video understanding through its mobile interface.
VITA-1.5~\cite{DBLP:journals/corr/abs-2501-01957} is the first open-source Multimodal Large Language Model capable of jointly processing video, image, text, and audio inputs while supporting advanced interactive experiences.
Built on Mixtral 8×7B with an expanded Chinese vocabulary and bilingual instruction tuning, VITA-1.5 enables real-time vision and speech interaction and can be deployed on the web for user access.

%----------------------------------
\subsection{Assistive Tools for the Visually Impaired}
%----------------------------------

Currently, both industry and academia have developed a wide range of assistive devices and software for severely visually impaired individuals, which can be broadly categorized into four main types.
Navigation aids, such as the WeWALK smart cane~\cite{wewalk} and the Sunu smart band~\cite{nfb_sunu_band}, help users detect obstacles using technologies like ultrasonic sensors or cameras, and provide feedback through vibrations or voice alerts. 
Image recognition applications, including the OrCam MyEye device~\cite{OrCam} and the ``Be My Eyes'' volunteer assistant app~\cite{BeMyEyes}, leverage optical character recognition to read text, describe simple scenes, or connect users with sighted volunteers via video calls for real-time assistance. 
Daily life assistive devices, such as the Victor Reader Stream digital audiobook player~\cite{HumanWare}, enable visually impaired users to access books and other reading materials in accessible formats.
Lastly, software-based tools, like the JAWS screen reader~\cite{freedomscientific_jaws}, allow visually impaired users to operate computers through speech output, making digital content more accessible.
However, these products cannot understand environmental semantics or contextual tasks. 
They cannot guide severely visually impaired users through daily goals like ``find this book'' or answer questions such as ``what is in front of me'' that require natural interactive communication.

With the emergence of LLMs and VLMs, many researchers leverage their powerful understanding capabilities to provide better support for visually impaired individuals.
Kuzdeuov et al.~\cite{DBLP:conf/icaiic/KuzdeuovMNKV24} develop a mobile application specifically designed for visually impaired users, allowing them to interact with ChatGPT through natural speech.
This enables easier access to information and task completion, helping users live and learn more independently.
Zhang et al.~\cite{DBLP:conf/chi/Zhang24a} design a novel interactive system for the visually impaired that combines image semantic segmentation, language models, and haptic feedback.
This system allows users to ``see'' objects and semantic information in images through voice and touch.
Zhao et al.~\cite{DBLP:journals/corr/abs-2402-01735} propose a task framework named VIALM (Visually Impaired Assistance with Large Models), focusing on the application of large models in visual impairments assistance.
However, real-world environments are dynamic. 
Relying solely on static image understanding is insufficient to meet the needs of visually impaired individuals for continuous perception in real-life situations.
There is still a lack of in-depth exploration of continuous visual inputs such as video, which is the central focus of this study.

\begin{table*}[t]
\centering
\caption{Questionnaire on mobility, household, and social tasks. Symbols: \LEFTcircle\ denotes single-choice questions, \CIRCLE\ denotes multiple-choice questions.}
\label{tab:questionnaire}
\begin{tabularx}{\textwidth}{c c p{0.28\textwidth} X}
\toprule
\textbf{Domain} & \textbf{Type} & \textbf{Question} & \textbf{Options} \\
\midrule

\multirow{3}{*}{\parbox{2.0cm}{\centering Basic \\ Skills}} 
& \LEFTcircle & How difficult do you find it to orient yourself in unfamiliar environments? & 
a) Very difficult \quad b) Mostly difficult \quad c) Moderate \quad d) Slightly difficult \quad e) Not difficult at all \\
\cmidrule(l){2-4}
& \LEFTcircle & Do you feel comfortable walking independently in familiar environments? & 
a) Very dissatisfied \quad b) Dissatisfied \quad c) Neutral \quad d) Satisfied \quad e) Very satisfied \\
\cmidrule(l){2-4}
& \CIRCLE & What problems do you commonly face when walking with a cane? & 
a) Incomplete/invisible accessible paths \quad b) Too many obstacles (cars, objects) \quad c) Difficulty judging terrain changes (steps, slopes) \quad d) Lack of cane training \quad e) Others \\
\midrule

\multirow{3}{*}{\parbox{2.0cm}{\centering Home Life \\ Tasks}} 
& \CIRCLE & What household tasks do you most often need help with? & 
a) Cleaning \quad b) Organizing/storing items \quad c) Cooking/preparing food \quad d) Washing clothes \quad e) Others \\
\cmidrule(l){2-4}
& \CIRCLE & What are your main means of leisure and entertainment? & 
a) Listening to radio/audiobooks \quad b) Using smartphone/computer apps (screen reader, voice) \quad c) Chatting with family/friends \quad d) Community activities \quad e) Others \\
\cmidrule(l){2-4}
& \CIRCLE & If given more training/support, in which areas would you like to improve your independence? & 
a) Household management \quad b) Using tech products (phone, computer, smart speaker) \quad c) Time management/planning \quad d) Leisure/recreation skills \quad e) Others \\
\midrule

\multirow{4}{*}{\parbox{2.0cm}{\centering Social Life\\ Tasks}} 
& \CIRCLE & What social barriers do you most commonly encounter? & 
a) Difficult to reach venues \quad b) Lack of accessible facilities \quad c) Inaccessible info (no voice prompts) \quad d) Communication barriers \quad e) Lack of companionship \quad f) Others \\
\cmidrule(l){2-4}
& \LEFTcircle & How difficult is walking during social activities for you? & 
a) Very difficult \quad b) Mostly difficult \quad c) Moderate \quad d) Slightly difficult \quad e) Not difficult at all \\
\cmidrule(l){2-4}
& \CIRCLE & What are the main difficulties you face when using public transportation? & 
a) Difficult to find station/entrance \quad b) Difficult to get transfer/arrival info \quad c) Inconvenient boarding/alighting \quad d) Lack of support from staff \quad e) Others \\
\cmidrule(l){2-4}
& \LEFTcircle & How would you assess your mobility in daily social life? & 
a) Adequate, meets my needs \quad b) Fair, with room for improvement \quad c) Inadequate, often restricted \\
\bottomrule
\end{tabularx}
\end{table*}

\begin{figure*}[t]
\centering
\captionsetup[subfigure]{font=small,justification=centering}
% -------- Row 1 --------
\begin{subfigure}[t]{\textwidth}
  \centering
  \includegraphics[width=\textwidth]{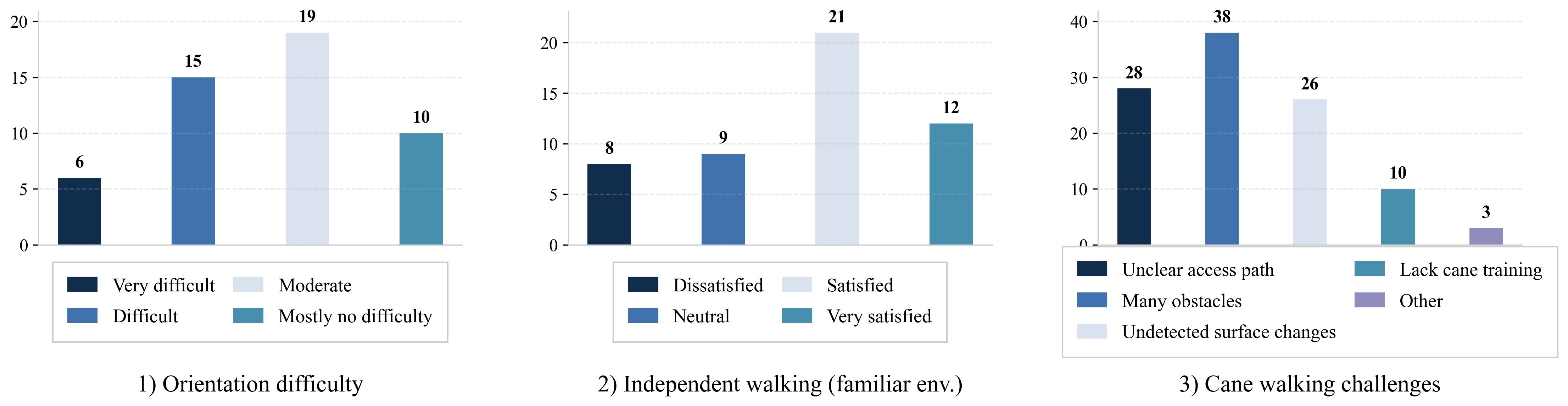}
    \caption{Basic skills.}
  \label{fig:basic}
\end{subfigure}\hfill

\vspace{0.6em}

% -------- Row 2 --------
\begin{subfigure}[t]{\textwidth}
  \centering
  \includegraphics[width=\textwidth]{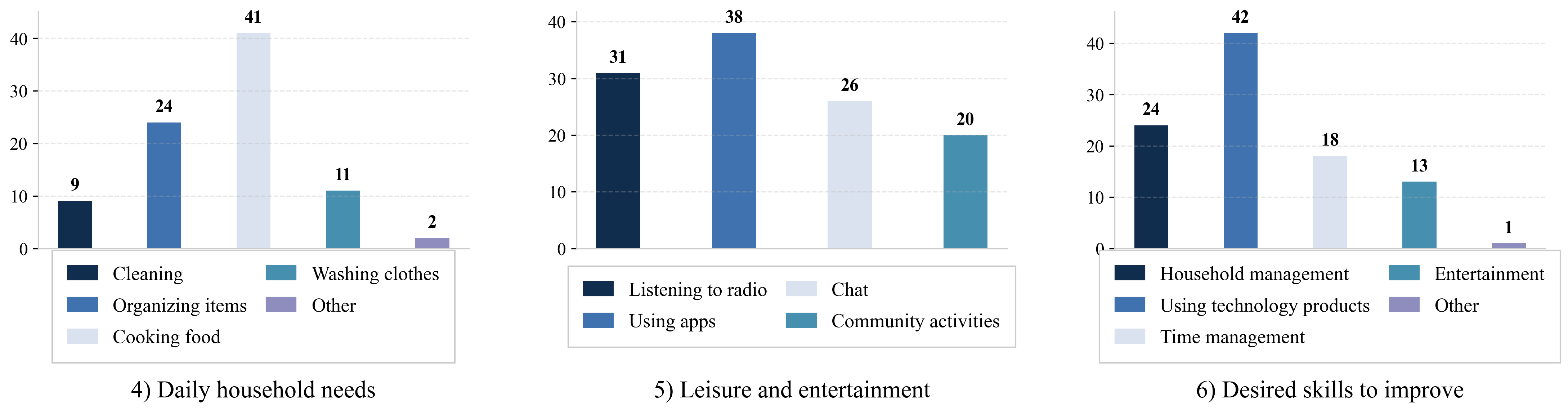}
    \caption{Home Life Tasks.}
  \label{fig:home}
\end{subfigure}\hfill

\vspace{0.6em}

% -------- Row 3 --------
\begin{subfigure}[t]{\textwidth}
  \centering
  \includegraphics[width=\textwidth]{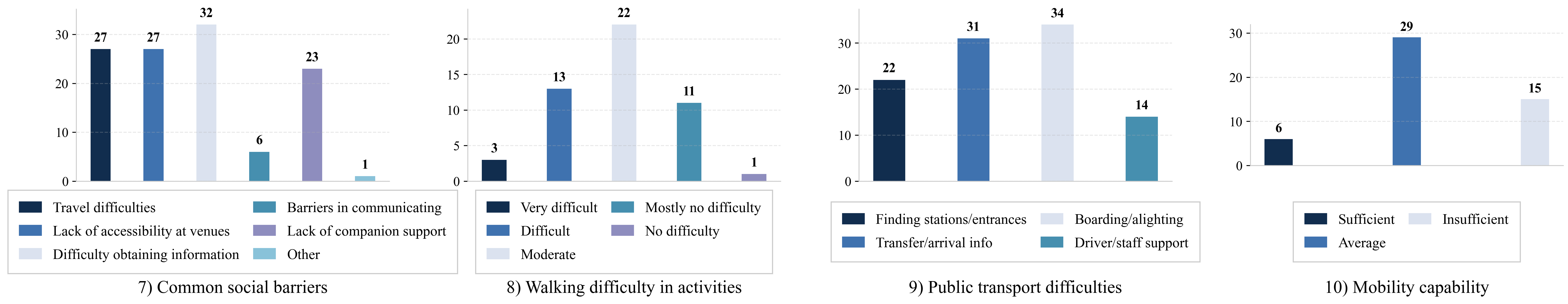}
    \caption{Social Life Tasks.}
  \label{fig:social}
\end{subfigure}\hfill
\vspace{0.6em}

\caption{Illustrations for questionnaire items.}
\label{fig:questionnaire-all}
\end{figure*}

\section{User Survey}
To ensure that our evaluation of VideoLLMs accurately reflects the core needs of the individual with visual impairments, we design and conduct a questionnaire together with short semi-structured interviews.
Participants are directly involved in the construction of the evaluation benchmark. The questionnaire covers three main themes, including daily mobility, home life activities, and social participation (as shown~\Cref{tab:questionnaire}), which grounds the design of \bench in user needs and real-world challenges.

\subsection{Set Up}

\mypara{Participants}
We recruit a total of 50 adults with visual impairments, including 24 men (48\%), 23 women (46\%), and 3 non-binary or other individuals (6\%).
In terms of visual status, 24 participants are blind (48\%) and 26 are low-vision (52\%).
All participants provide informed consent after being fully informed of the study purpose and data use.
% ~\footnote{Ethical approval is obtained from the IRB, approval number xxx.}

\mypara{Tools and Procedure}
The questionnaire items are grouped into basic skills, household tasks, and social tasks, using single-choice or multiple-choice formats. 
Optional semi-structured interviews follow the questionnaire, focusing on recent challenges and expectations for AI assistance.
The questionnaire is distributed to real blind communities and online platforms for blind communities, such as Reddit.  
In addition, two open-ended questions are analyzed through inductive thematic analysis, and word clouds are created as complementary visualizations (as shown~\Cref{fig:wordclouds} for more details). 
Synonyms are merged during processing, for example, ``AI assistant'', ``voice assistant'', and ``smart life assistant'' are combined into “AI/voice assistant.''

\subsection{Results}

\mypara{Basic Skills}
In the area of basic skills, participants report significant difficulties as shown in~\Cref{fig:basic}.

\begin{enumerate}
    \item \textbf{Orientation ability.} In unfamiliar environments, most participants report at least moderate difficulty: 6 ``very difficult'', 15 ``difficult'', and 19 ``moderate'', while 10 report ``mostly no difficulty''.
    This underscores the need for VideoLLMs to recognize environmental features and provide reliable directional guidance.
    \item \textbf{Independent walking.} In familiar environments, most respondents express satisfaction with their ability to walk independently: 21 ``satisfied'' and 12 ``very satisfied'', with 9 ``neutral'' and 8 ``dissatisfied''. 
    Interview responses suggest this confidence depends heavily on environmental stability and drops quickly when obstacles or changes occur.
    \item \textbf{Challenges in cane walking.} The most frequent difficulties selected are ``many obstacles'' (38) and ``unclear access path'' (28), followed by ``undetected surface changes'' (26) , ``lack of cane training'' (10), and ``other'' (2).
\end{enumerate}
These findings highlight the importance of obstacle recognition, path guidance, and technique support in our benchmark tasks.

\mypara{Household Life Tasks}
In the area of household tasks, participants report significant difficulties as shown in~\Cref{fig:home}.

\begin{enumerate}
    \item \textbf{Daily household needs.}
    Cooking or meal preparation is selected 41 times, organizing or storing items 24 times, doing laundry 11 times, cleaning 9 times, and other activities 2 times.
    The results indicate that the main challenges in household life center on meal preparation and item management.
    \item \textbf{Leisure and entertainment activities.}
    Using mobile or computer applications is selected 38 times, listening to audio content 31 times, chatting with family or friends 26 times, and participating in community activities 20 times. These results suggest that mobile devices have become the primary entry point for information access and leisure.
    \item \textbf{Desired skills to improve.}
    The use of digital devices is selected 42 times, household management 24 times, time management or planning 18 times, leisure skills 13 times, and other activities 1 time.
\end{enumerate}

These findings highlight the importance of household item management, mobile-based interaction, and digital device usage in our benchmark tasks.

\mypara{Social Life Tasks}
In the area of social life tasks, participants report significant difficulties as shown in~\Cref{fig:social}.
\begin{enumerate}
    \item \textbf{Common social barriers.} Reported frequencies are: travel difficulties 27, insufficient accessibility of venues 27, difficulties in accessing information 32, lack of companionship support 23, communication barriers 6, and other barriers 1. Overall, information accessibility and the lack of accessible facilities in public spaces emerge as the most prominent issues, while the results also reflect limited companionship support and social resources.
    \item \textbf{Walking difficulties in social activities.}
    Very difficult 3, difficult 13, moderate 22, mostly no difficulty 11, no difficulty 1.
    Most participants report moderate or lighter levels of difficulty, suggesting that environmental conditions often play a more decisive role than personal walking ability.
    \item \textbf{Main difficulties in public transportation.}
    Locating stops or entrances 22, transfer and arrival information 31, getting on or off vehicles 34, support from drivers or staff 14.
    The results indicate that the main obstacles in travel are concentrated in station recognition, access to transportation information, and the process of boarding and alighting.
    \item \textbf{Self-assessment of mobility in daily social life.}
    Sufficient 6, average 29, insufficient 15.
    The majority of participants rate themselves as ``average'', though a considerable proportion still consider their ability insufficient.
\end{enumerate}

These findings highlight the importance of information access, accessible venues, and public transport support in our benchmark tasks.

\begin{figure}[h!]
    \centering
    \begin{subfigure}[t]{0.38\textwidth}
        \centering
        \includegraphics[width=\textwidth]{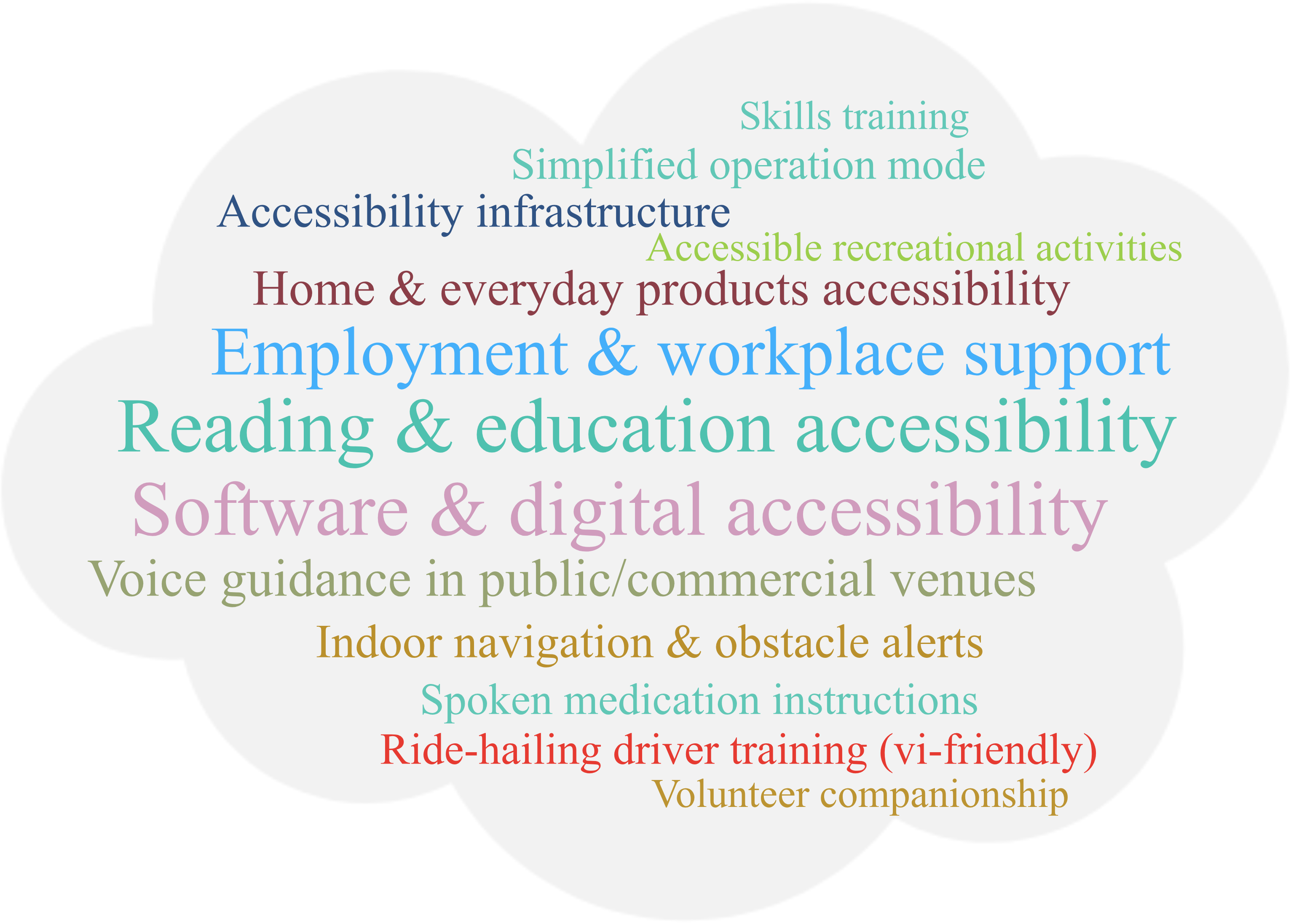}
        \caption{Word cloud results for Q1: areas to improve beyond mobility.}
        \label{fig:overviewQ1}
    \end{subfigure}%
    \hfill
    \begin{subfigure}[t]{0.48\textwidth}
        \centering
        \includegraphics[width=\textwidth]{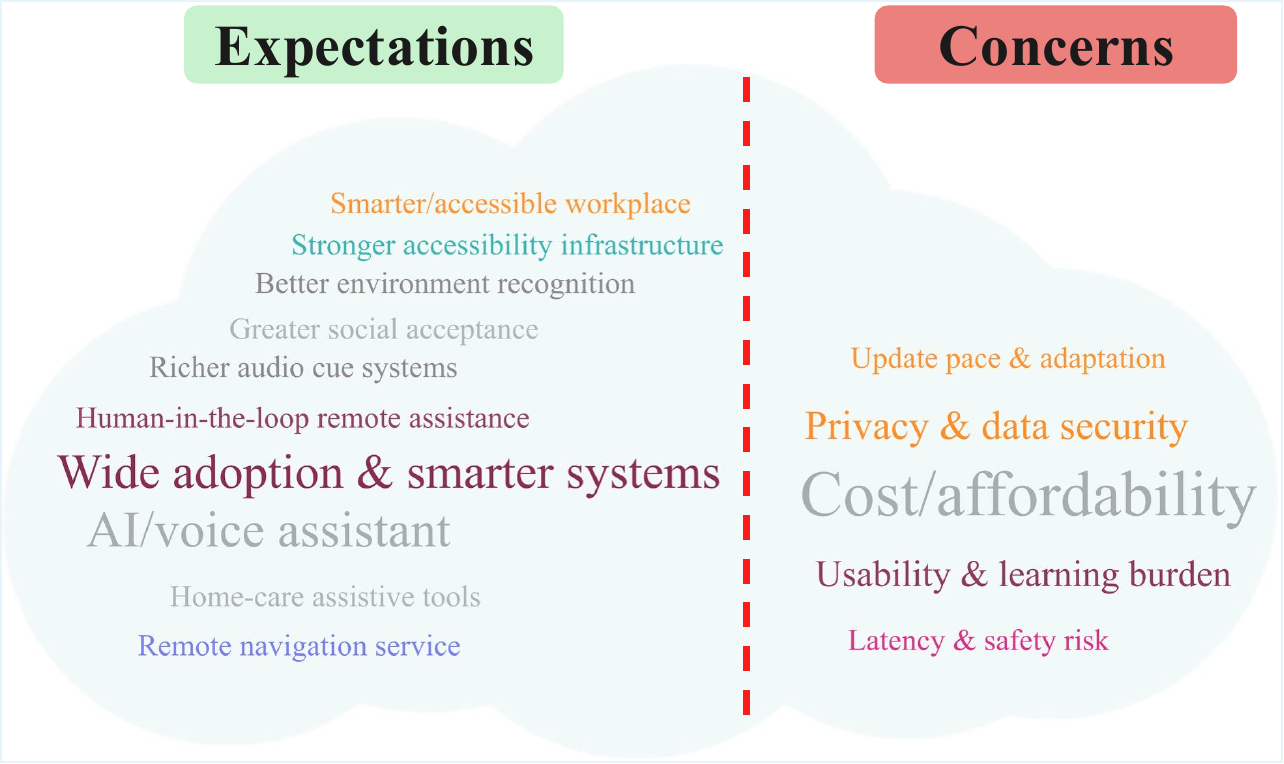}
        \caption{Word cloud results for Q2: expectations and concerns for future assistive technologies.}
        \label{fig:overviewQ2}
    \end{subfigure}
    \caption{Word clouds for the user survey's open-ended results.}
    \label{fig:wordclouds}
\end{figure}

\mypara{Open-ended Results}
\begin{enumerate}
    \item \textbf{Q1: Areas to improve beyond mobility (n=33).}
    The most frequently mentioned topics are reading and education accessibility (6/33) and software or digital accessibility (6/33), followed by employment and workplace support (4/33) and voice guidance in public or commercial venues (4/33).
    Other mentions include home and everyday product accessibility (3/33), signage and information presentation (3/33), accessibility infrastructure (2/33), indoor navigation and obstacle alerts (2/33), and several single mentions (1/33 each): spoken medication instructions, ride-hailing driver training, simplified operation modes, volunteer companionship, accessible recreational activities, and skills training (as shown~\Cref{fig:overviewQ1}).
    
    Overall, the responses converge on two clusters: information and content accessibility (reading, software, signage, simplified modes) and contextual guidance for participation (voice guidance, workplace support).
    At the same time, the long tail of single mentions reveals diverse and highly individualized needs.
    \item \textbf{Q2: Expectations and concerns for future assistive technologies (n=23).}
    Positive expectations highlight wide adoption and smarter systems (5/23) and AI or voice assistants (4/23), with single mentions of human-in-the-loop remote assistance, remote navigation, home-care tools, greater social acceptance, better environment recognition, stronger accessibility infrastructure, richer audio cue systems, and accessible workplaces. Concerns mainly focus on cost and affordability (3/23), privacy and data security (2/23), and usability or learning burden (2/23), with isolated mentions of latency and safety risks (1/23) and the pace of updates and adaptation (1/23)(as shown~\Cref{fig:overviewQ2}).
    
    Users envision assistive systems that are both intelligent and ubiquitous, but their primary worries concern affordability, trust in data practices, and the ease of learning. This combination suggests a strong demand for mainstream, low-barrier, and privacy-conscious design.
\end{enumerate}

\begin{table*}[t]
\centering
\caption{Basic Skills (Tool: Cane).}
\label{table:basic_skills}
\setlength{\tabcolsep}{6pt}
\renewcommand{\arraystretch}{1.25}
\begin{tabular}{|
  >{\centering\arraybackslash}p{0.15\textwidth} |
  >{\raggedright\arraybackslash}p{0.19\textwidth} |
  >{\raggedright\arraybackslash}p{0.16\textwidth} |
  >{\raggedright\arraybackslash}p{0.31\textwidth} |
  >{\raggedright\arraybackslash}p{0.05\textwidth} |
}
\Xhline{1.2pt}
\multicolumn{1}{|c|}{\textbf{Image}} &
\multicolumn{1}{c|}{\textbf{Task}} &
\multicolumn{1}{c|}{\textbf{Goal}} &
\multicolumn{1}{c|}{\textbf{Instruction}} &
\multicolumn{1}{c|}{\makecell[c]{\textbf{Location}\\\textbf{\& Tool}}} 
 \\

\Xhline{1.2pt}

% ===== Row 1 =====
\begin{minipage}[t]{\linewidth}\vspace{1pt}\centering
  \includegraphics[width=\linewidth]{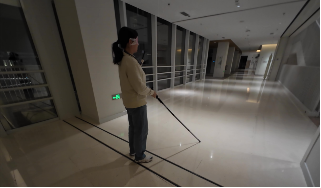}
\end{minipage} &
\begin{itemize}[leftmargin=*, label={},nosep]
  \item \textbf{Orientation Skills}
  \item(a) Ask environment
  \item(b) Describe obstacles
  \item(c) Determine facing direction
\end{itemize} &
\begin{itemize}[leftmargin=*, label={},nosep]
  \item Evaluates the model’s ability to help users determine direction in complex environments.
\end{itemize}
 &
\begin{itemize}[leftmargin=*,label={},nosep]
  \item(a) I am a blind person. Please describe the environment in front of me.
  \item(b) Could you please tell me if there are any obstacles in front of me?
  \item(c) Could you tell me which direction I’m facing to?
\end{itemize} &
\begin{itemize}[leftmargin=*, label={},nosep]
  \item Outdoor / Yes 
\end{itemize}\\
\hline
% ===== Row 2 =====
\begin{minipage}[t]{\linewidth}\vspace{1pt}\centering
  \includegraphics[width=\linewidth]{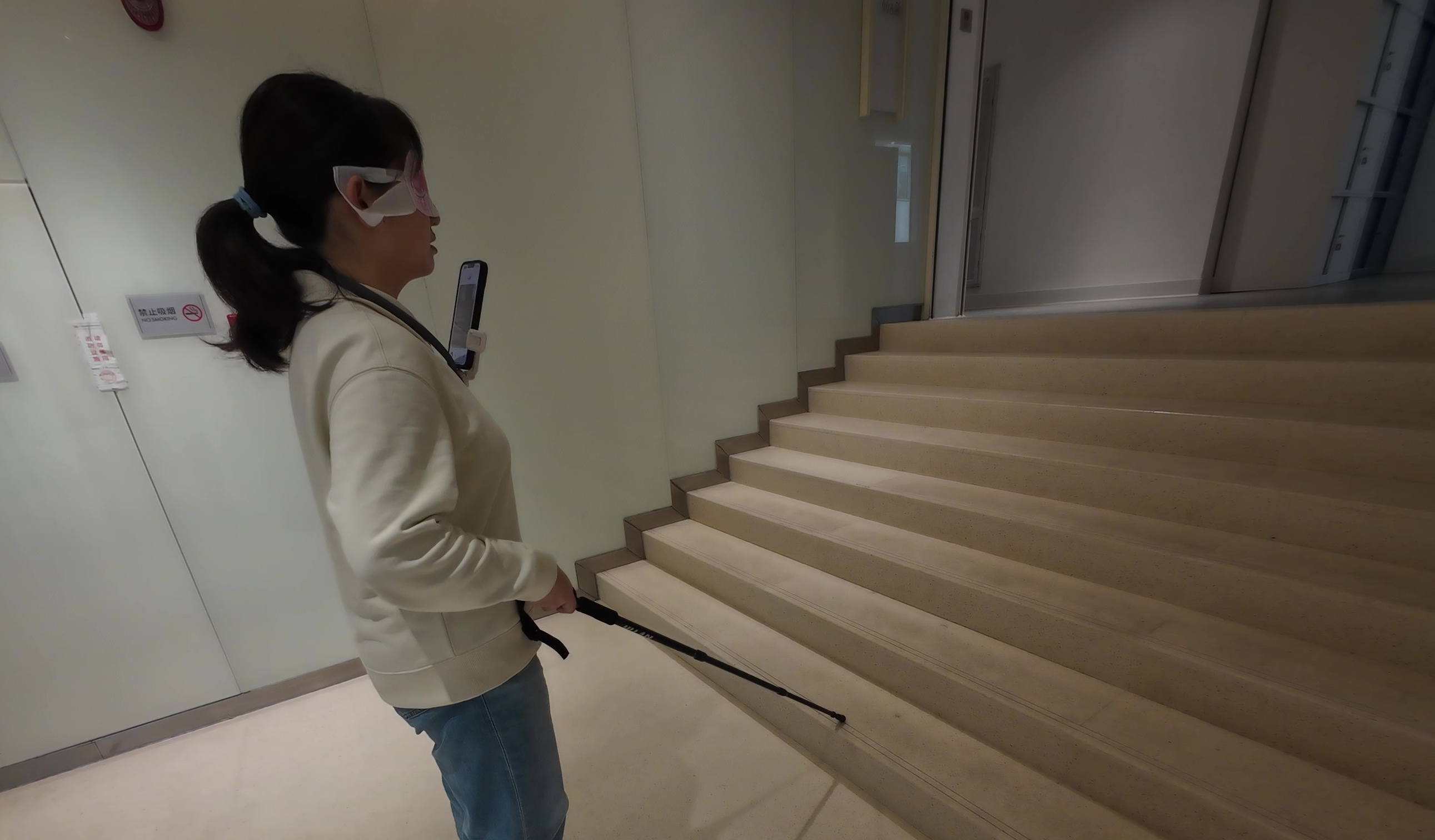}
\end{minipage} &
\begin{itemize}[leftmargin=*, label={}, nosep]
  \item \textbf{Guided Walking}
  \item (a) Ask staircase details
  \item (b) Guide ascent
  \item (c) Go upstairs
  \item (d) Make it upstairs
\end{itemize} &
\begin{itemize}[leftmargin=*, label={}, nosep]
  \item Tests if the model can safely guide stair climbing using real-time cane feedback.
\end{itemize} &
\begin{itemize}[leftmargin=*, label={}, nosep]
  \item (a) I am a blind person. Please tell me what is in front of me.
  \item (b) Please guide me step by step as I climb the stairs.
  \item (c) I am starting to go up the stairs now. Please help me to do it.
  \item (d) I have reached the top of the stairs. Can you describe the area ahead?
\end{itemize} &
\begin{itemize}[leftmargin=*, label={}, nosep]
  \item Outdoor / Yes
\end{itemize} \\
\hline

% ===== Row 3 =====
\begin{minipage}[t]{\linewidth}\vspace{1pt}\centering
  \includegraphics[width=\linewidth]{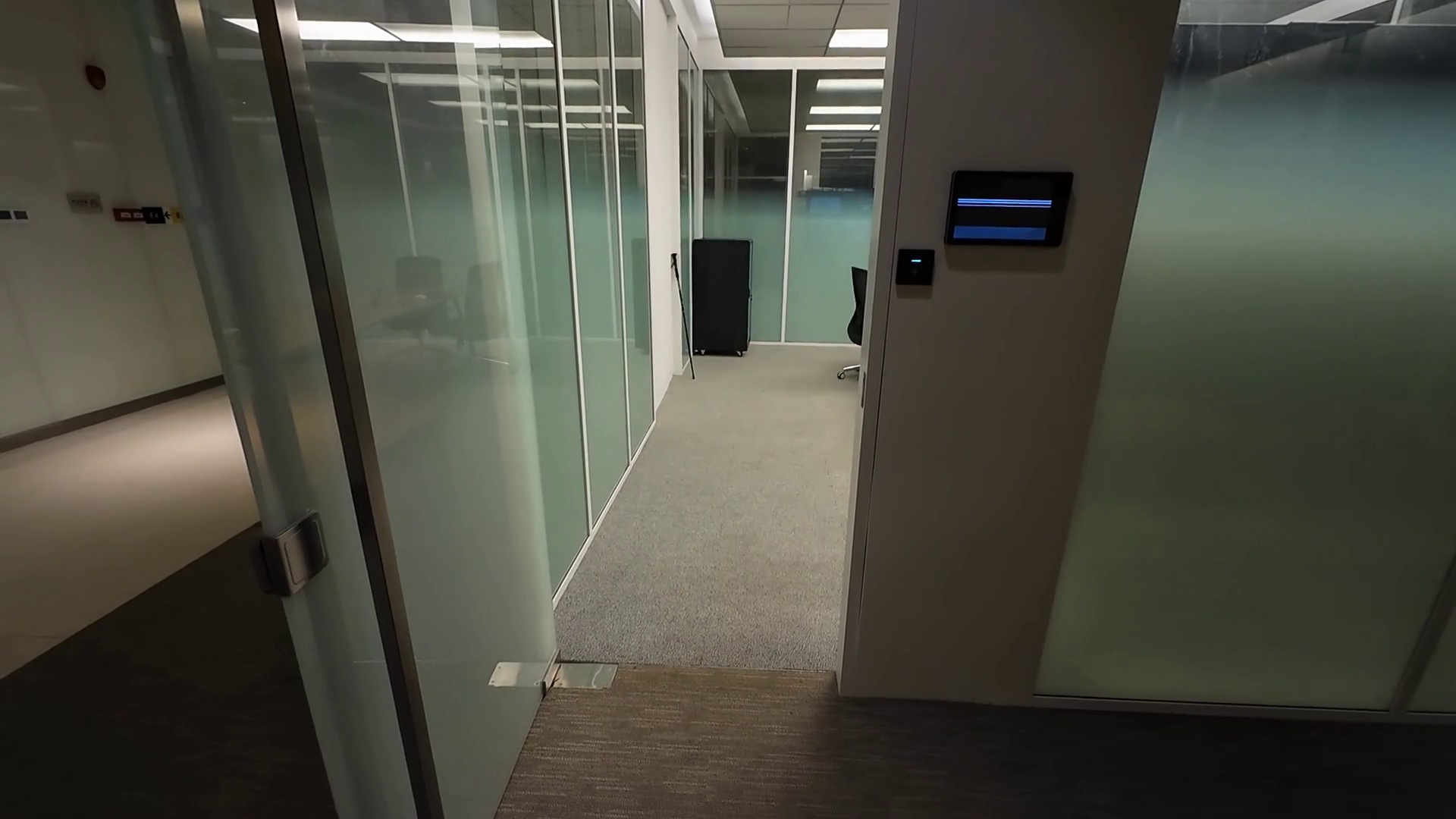}
\end{minipage} &
\begin{itemize}[leftmargin=*, label={}, nosep]
  \item \textbf{Independent Walking}
  \item (a) Ask room layout
  \item (b) Identify entry conditions
  \item (c) Locate lost object
\end{itemize} &
\begin{itemize}[leftmargin=*, label={}, nosep]
  \item Assesses whether the model can direct a user into a room without tools.
\end{itemize} &
\begin{itemize}[leftmargin=*, label={}, nosep]
  \item (a) I am a blind person. Can you describe the room in front of me?
  \item (b) What should I do if I would like to go into this room?
  \item (c) I have lost my cane in the room. Can you help me to find it?
\end{itemize} &
\begin{itemize}[leftmargin=*, label={}, nosep]
  \item Indoor / No
\end{itemize} \\
\hline

% ===== Row 4 =====
\begin{minipage}[t]{\linewidth}\vspace{1pt}\centering
  \includegraphics[width=\linewidth]{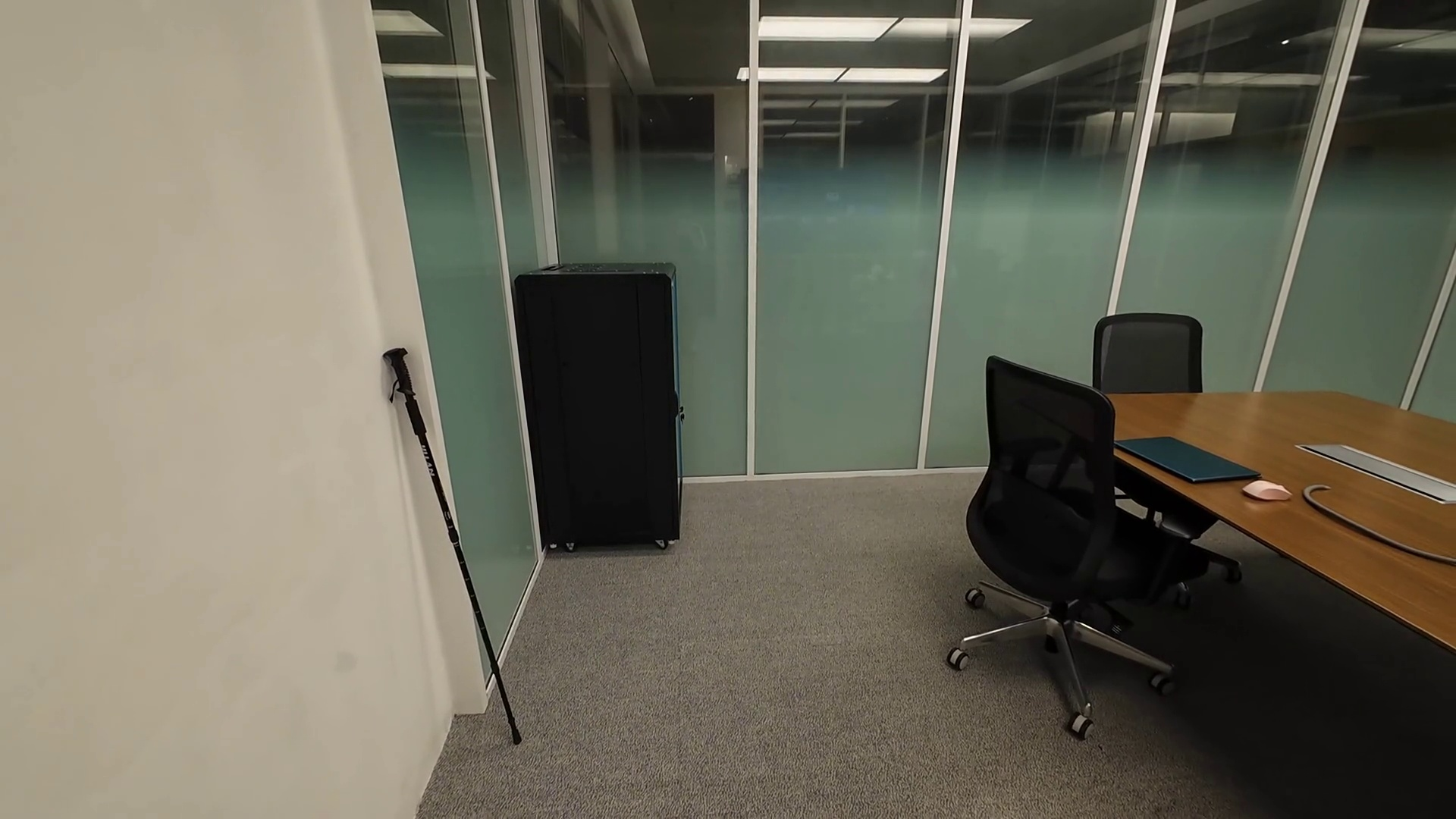}
\end{minipage} &
\begin{itemize}[leftmargin=*, label={}, nosep]
  \item \textbf{Cane Techniques}
  \item (a) Navigate along wall
  \item (b) Retrieve cane
  \item (c) Guide to seating
\end{itemize} &
\begin{itemize}[leftmargin=*, label={}, nosep]
  \item Evaluates the model’s ability to recognize and guide proper cane use.
\end{itemize} &
\begin{itemize}[leftmargin=*, label={}, nosep]
  \item (a) I am a blind person. Please guide me to follow the wall.
  \item (b) I need to find and pick up my cane, can you help me to find it?
  \item (c) I have found my cane, please guide me to the nearest seat.
\end{itemize} &
\begin{itemize}[leftmargin=*, label={}, nosep]
  \item Indoor / No
\end{itemize} \\

\Xhline{1.2pt}
\end{tabular}
\end{table*}
\begin{table*}[t]
\centering
\caption{Home Life Tasks (Tool: Cane).}
\label{table:home_life}
\setlength{\tabcolsep}{6pt}
\renewcommand{\arraystretch}{1.25}

\begin{tabular}{|
  >{\centering\arraybackslash}p{0.15\textwidth} |
  >{\raggedright\arraybackslash}p{0.19\textwidth} |
  >{\raggedright\arraybackslash}p{0.16\textwidth} |
  >{\raggedright\arraybackslash}p{0.31\textwidth} |
  >{\raggedright\arraybackslash}p{0.05\textwidth} |
}
\Xhline{1.2pt}
\multicolumn{1}{|c|}{\textbf{Image}} &
\multicolumn{1}{c|}{\textbf{Task}} &
\multicolumn{1}{c|}{\textbf{Goal}} &
\multicolumn{1}{c|}{\textbf{Instruction}} &
\multicolumn{1}{c|}{\makecell[c]{\textbf{Location}\\\textbf{\& Tool}}} 
 \\

\Xhline{1.2pt}
% ===== Row 1: Housework =====
\begin{minipage}[t]{\linewidth}\vspace{1pt}\centering
  \includegraphics[width=\linewidth]{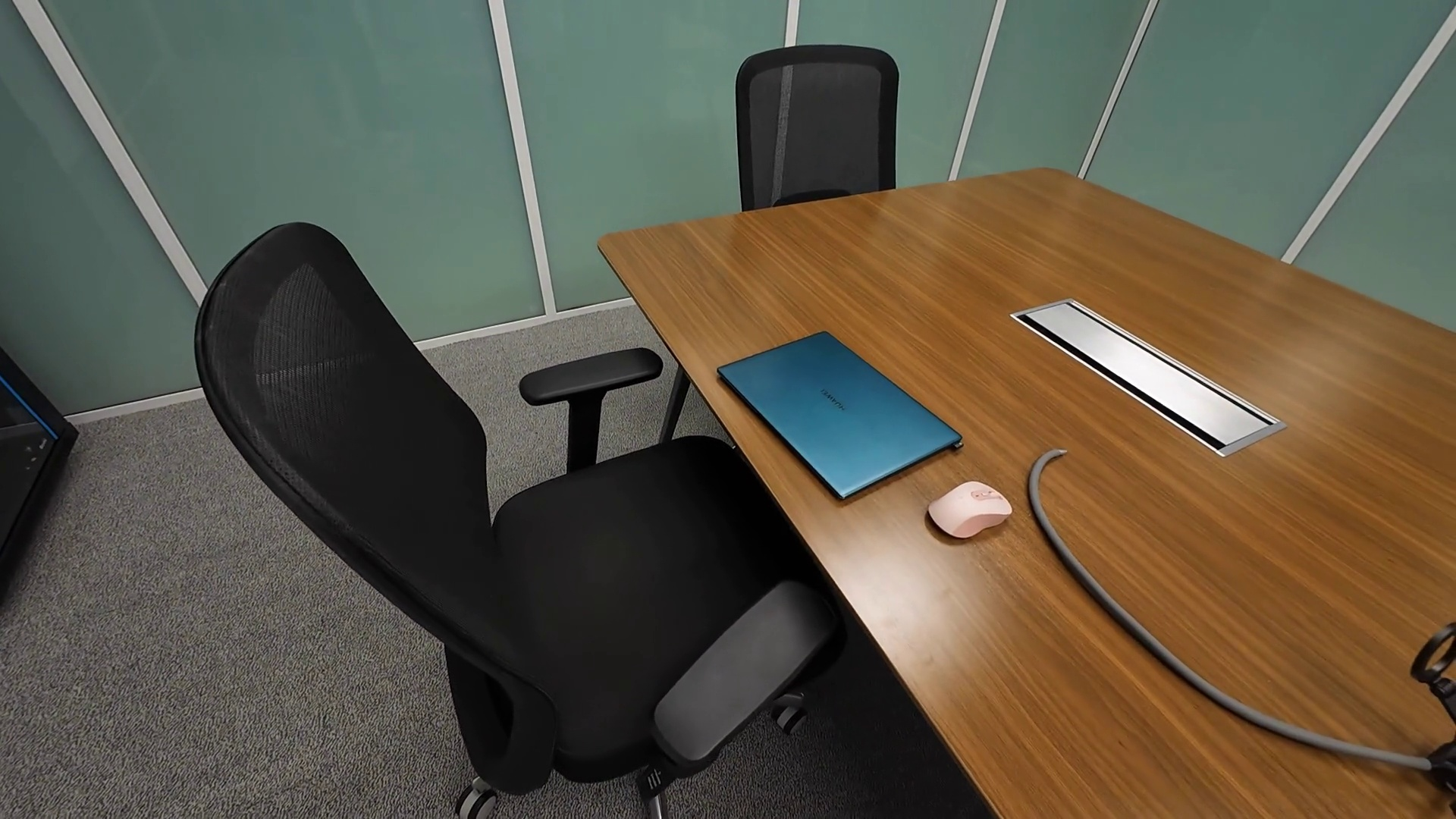}
\end{minipage} &
\begin{itemize}[leftmargin=*,label={},nosep]
  \item \textbf{Housework}
  \item (a) Find object ahead
  \item (b) Determine position
  \item (c) Identify the property
\end{itemize} &
\begin{itemize}[leftmargin=*,label={},nosep]
  \item Tests the model’s ability to identify object locations and attributes at home.
\end{itemize} &
\begin{itemize}[leftmargin=*,label={},nosep]
  \item (a) I am a blind person. Please help me identify what objects are in front of me.
  \item (b) Can you tell me exactly where this object is located relative to me?
  \item (c) What kind of object is it?
\end{itemize} &
\begin{itemize}[leftmargin=*, label={}, nosep]
  \item Indoor / No
\end{itemize} \\
\hline

\begin{minipage}[t]{\linewidth}\vspace{1pt}\centering
  \includegraphics[width=\linewidth]{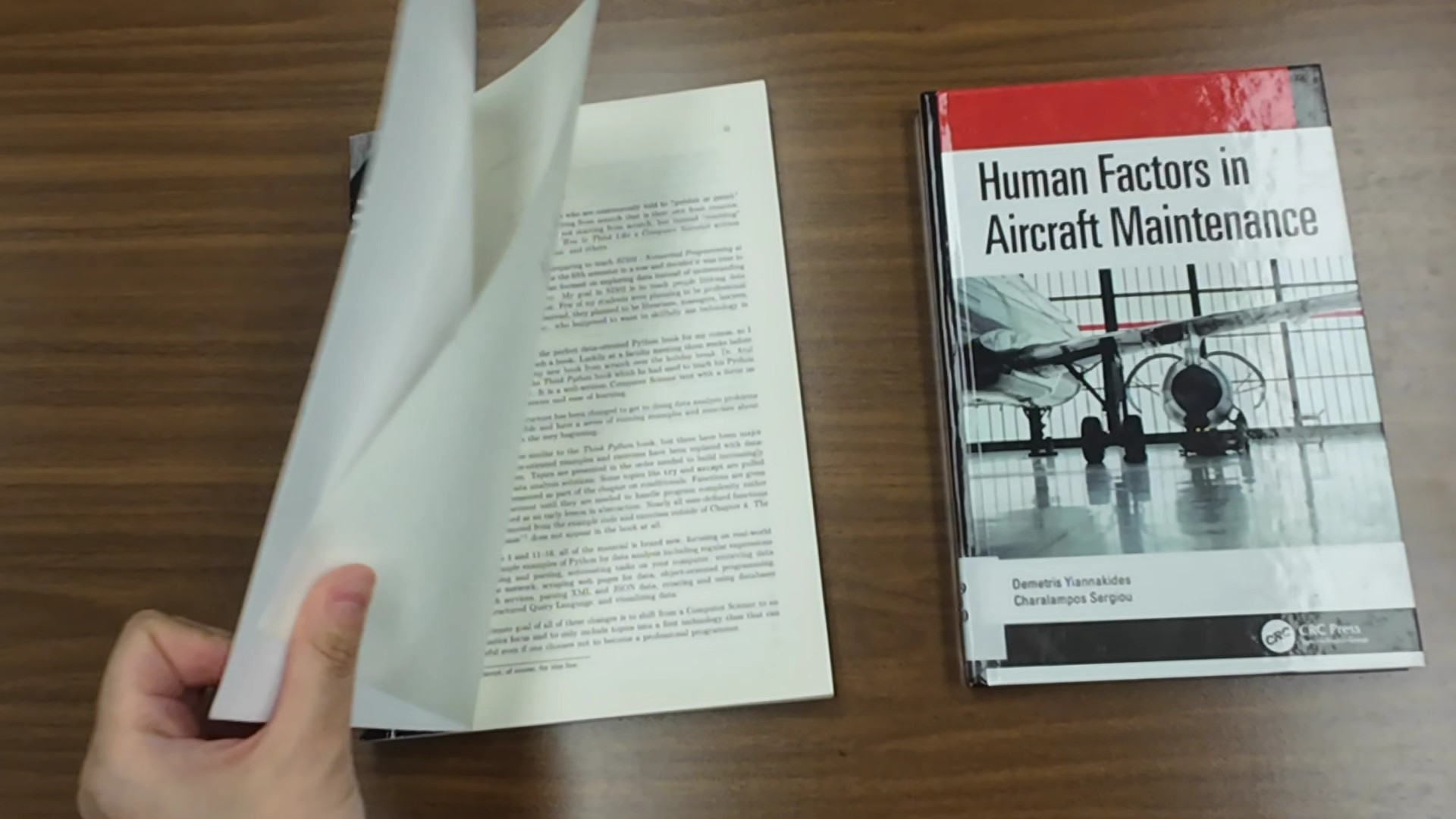}
\end{minipage} &
\begin{itemize}[leftmargin=*,label={},nosep]
  \item \textbf{Leisure and Recreation}
  \item (a) Identify the books
  \item (b) Turn the pages
  \item (c) Read the table of contents
\end{itemize} &
\begin{itemize}[leftmargin=*,label={},nosep]
  \item Assesses the model’s ability to help access printed books through content recognition.
\end{itemize} &
\begin{itemize}[leftmargin=*,label={},nosep]
  \item (a) I am a blind person. Can you help me identify which book is in front of me?
  \item (b) Please guide me on how to turn the pages of this book.
  \item (c) Can you tell me how to find the page of the table of contents?
\end{itemize} &
\begin{itemize}[leftmargin=*, label={}, nosep]
  \item Indoor / No
\end{itemize} \\
\hline
\Xhline{1.2pt}
\end{tabular}
\end{table*}
\begin{table*}[t]
\centering
\caption{Social Life Tasks (Tool: Cane).}
\label{table:social_life}
\setlength{\tabcolsep}{6pt}
\renewcommand{\arraystretch}{1.25}

\begin{tabular}{|
  >{\centering\arraybackslash}p{0.15\textwidth} |
  >{\raggedright\arraybackslash}p{0.19\textwidth} |
  >{\raggedright\arraybackslash}p{0.16\textwidth} |
  >{\raggedright\arraybackslash}p{0.31\textwidth} |
  >{\raggedright\arraybackslash}p{0.05\textwidth} |
}
\Xhline{1.2pt}
\multicolumn{1}{|c|}{\textbf{Image}} &
\multicolumn{1}{c|}{\textbf{Task}} &
\multicolumn{1}{c|}{\textbf{Goal}} &
\multicolumn{1}{c|}{\textbf{Instruction}} &
\multicolumn{1}{c|}{\makecell[c]{\textbf{Location}\\\textbf{\& Tool}}} \\
\Xhline{1.2pt}

\begin{minipage}[t]{\linewidth}\vspace{1pt}\centering
  \includegraphics[width=\linewidth]{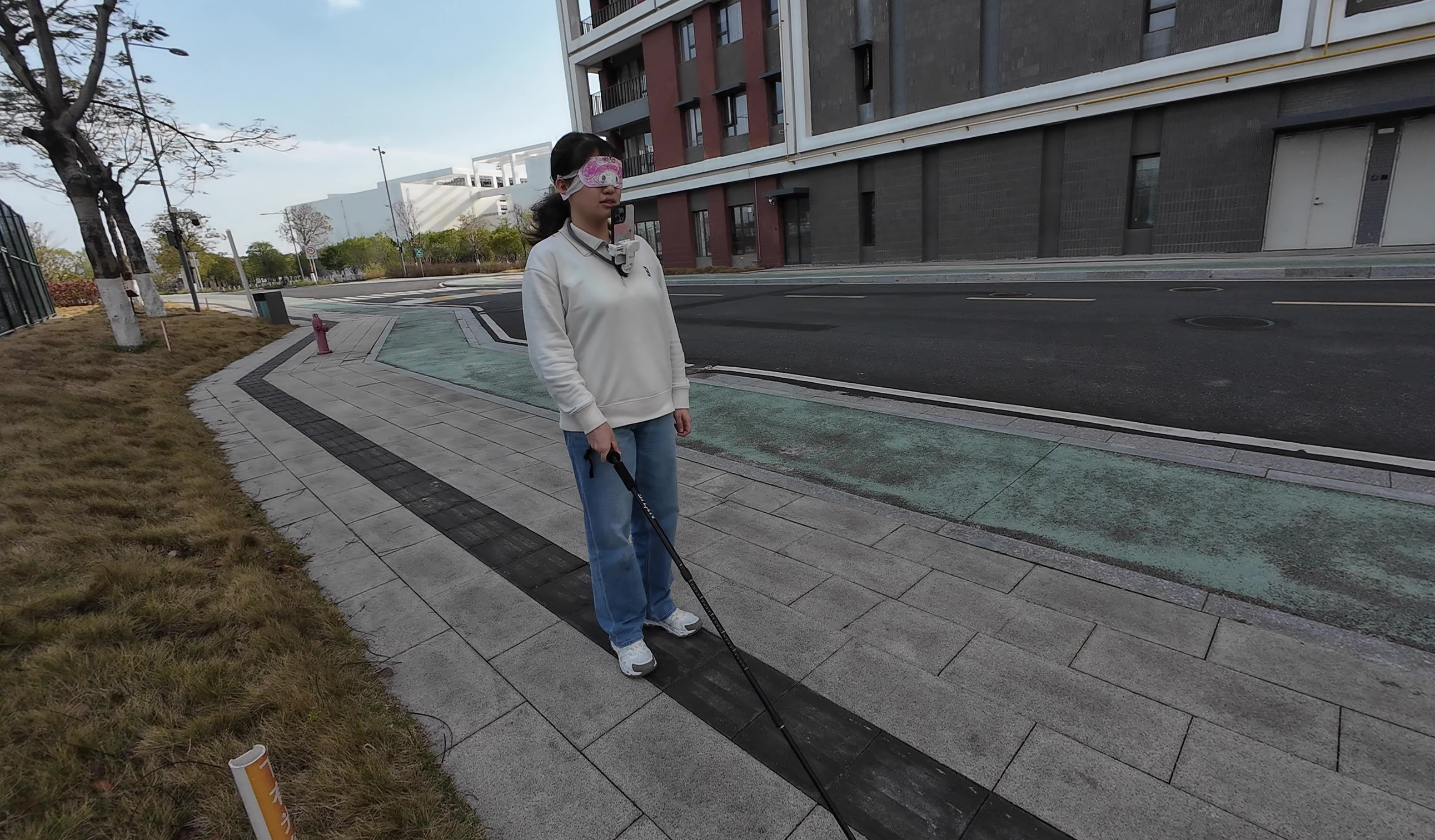} 
\end{minipage} &
\begin{itemize}[leftmargin=*,label={},nosep]
  \item \textbf{Road Walking}
  \item (a) Describe the environment
  \item (b) Walk on blind lane
  \item (c) Navigate by tapping a cane
\end{itemize} &
\begin{itemize}[leftmargin=*,label={},nosep]
  \item Evaluates the model’s ability to support safe navigation on roads using tactile paths.
\end{itemize} &
\begin{itemize}[leftmargin=*,label={},nosep]
  \item (a) I am a blind person. Please describe the surroundings around me.
  \item (b) I am on the blind lane, please guide me to walk safely along it.
  \item (c) As I tap my cane, can you help me figure out what’s around me and guide me forward?
\end{itemize} &
\begin{itemize}[leftmargin=*,label={},nosep]
  \item Outdoor / Yes
\end{itemize} \\
\hline

\begin{minipage}[t]{\linewidth}\vspace{1pt}\centering
  \includegraphics[width=\linewidth]{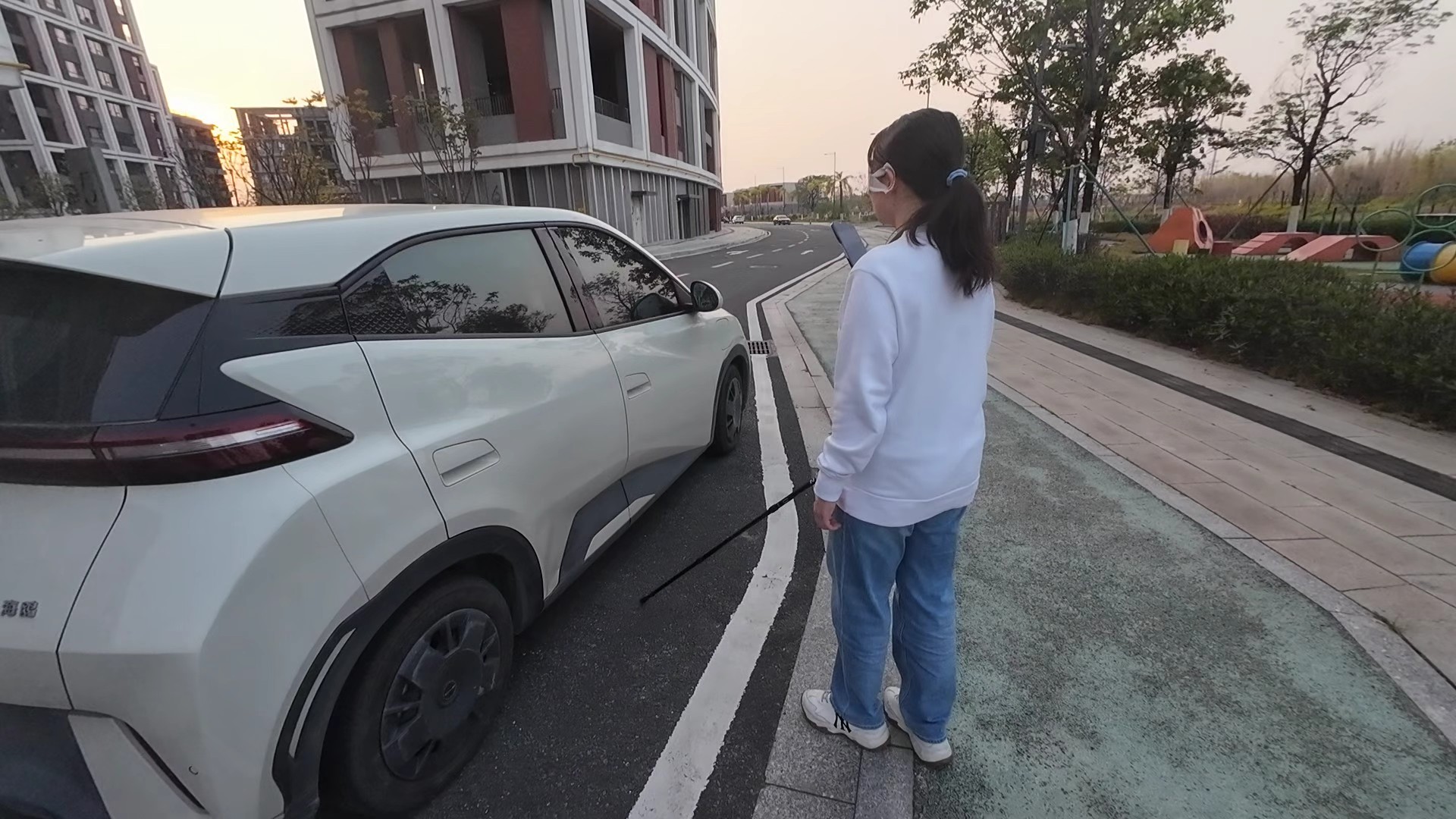} 
\end{minipage} &
\begin{itemize}[leftmargin=*,label={},nosep]
  \item \textbf{Transport}
  \item (a) Ask location of the car
  \item (b) Find the door
  \item (c) Open the door
  \item (d) Get on the car
\end{itemize} &
\begin{itemize}[leftmargin=*,label={},nosep]
  \item Tests the model’s ability to assist with vehicle identification and boarding.
\end{itemize} &
\begin{itemize}[leftmargin=*,label={},nosep]
  \item (a) I am a blind person. Can you tell me where the car is parked?
  \item (b) Please guide me to locate the car door.
  \item (c) Now I reach it, please help me open the door.
  \item (d) Please guide me to get into the car.
\end{itemize} &
\begin{itemize}[leftmargin=*,label={},nosep]
  \item Outdoor / Yes
\end{itemize} \\
\hline

\begin{minipage}[t]{\linewidth}\vspace{1pt}\centering
  \includegraphics[width=\linewidth]{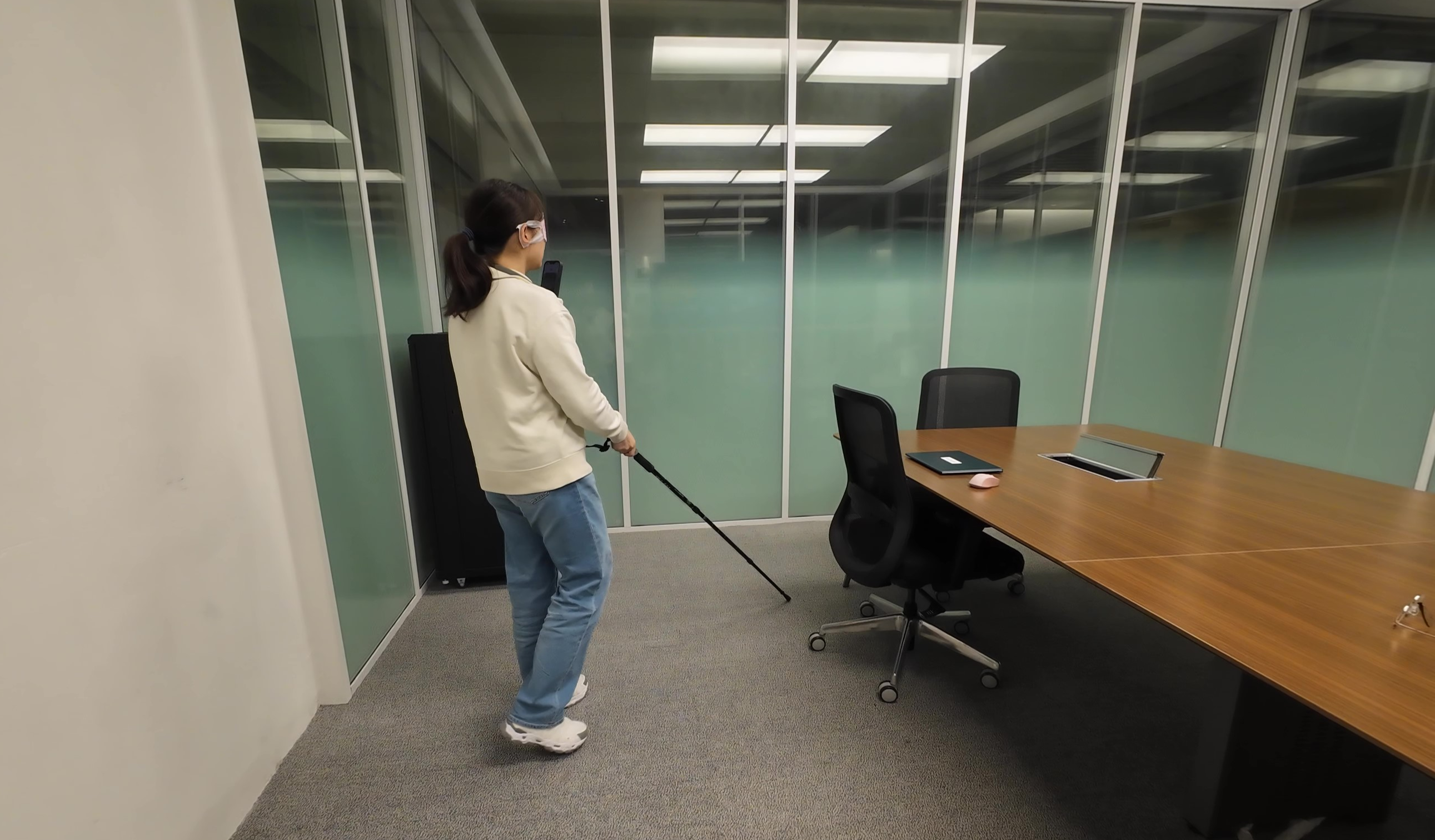} 
\end{minipage} &
\begin{itemize}[leftmargin=*,label={},nosep]
  \item \textbf{Reaching Destination}
  \item (a) Recognize environment
  \item (b) Make sure the location of chair
  \item (c) Judge the chair's position
  \item (d) Sit down
\end{itemize} &
\begin{itemize}[leftmargin=*,label={},nosep]
  \item Assesses whether the model can guide the user to complete final actions upon arrival.
\end{itemize} &
\begin{itemize}[leftmargin=*,label={},nosep]
  \item (a) I am a blind person. Can you help me recognize where I’ve arrived?
  \item (b) Is there a chair nearby? Please confirm its location.
  \item (c) Please help me to find the exact position of the chair.
  \item (d) Please guide me to sit down on the chair.
\end{itemize} &
\begin{itemize}[leftmargin=*,label={},nosep]
  \item Indoor / Yes
\end{itemize} \\
\Xhline{1.2pt}
\end{tabular}
\end{table*}

%----------------------------------
\section{Dataset}
%----------------------------------

In this paper, we construct two datasets. 
The first is an evaluation benchmark, \bench, which is used to assess the ability of VideoLLMs to assist severely visually impaired individuals in completing daily activities. 
The second dataset is a video-text paired dataset, \trainingdata, intended to support the model in proactively perceiving and alerting users to potential hazards in the environment.

%----------------------------------
\subsection{\bench}
%----------------------------------

Based on previously established standards~\cite{perkins_om_standards,cbra_orientation_training_2024}, we collaborate with visually impaired volunteers to design \bench, a benchmark for evaluating the ability of VideoLLMs to assist severely visually impaired individuals in completing daily activities.
The \bench is structured into three main modules, including Basic Skills, Home Life Tasks, and Social Life Tasks.

%----------------------------------
\subsubsection{Basic Skills} 
%----------------------------------

It focuses on evaluating whether VideoLLMs can assist the visually impaired individuals in performing basic daily mobility skills. 
The overall process, illustrated in the \Cref{table:basic_skills}, includes four tasks encompassing Orientation Skills, Guided Walking, Independent Walking, and Cane Techniques.

\mypara{Orientation Skills} 
In assisting visually impaired users in their daily activities, the ability to judge direction constitutes a key verification dimension of VideoLLMs' reliability. 
This module focuses on the model's ability to resolve directions in complex environments. 
The model should assist by analyzing the environment, describing obstacles and surface conditions, leveraging tactile paths, and guiding user orientation.

\mypara{Guided Walking} 
To assist visually impaired individuals in anticipating complex spatial layouts within unfamiliar environments, we design and evaluate a guided staircase-ascent procedure. 
While holding a cane, the user first communicates to the model a description of the surrounding environment and expresses the intention to climb the stairs. 
The model then instructs the user on how to probe each step’s height with the cane, issues real-time reminders throughout the ascent, and evaluates whether it is safe to proceed.

\mypara{Independent Walking} 
This task verifies whether the model can guide the user into a room without any tools.
This task evaluates the model’s spatial understanding and its ability to deliver clear and practical instructions.

\mypara{Cane Techniques} 
Individuals with visual impairments rely on the
cane for spatial probing during navigation, which necessitates frequent identification and retrieval of the cane’s location. 
This task evaluates the model’s ability to recognize and localize specialized assistive objects and assesses its effectiveness in guiding cane usage in real-world scenarios.

%----------------------------------
\subsubsection{Home Life Tasks}
%----------------------------------

These tasks evaluate the model’s ability to support individuals with visual impairments in completing household tasks and engaging in reading activities within a home environment, aiming to enhance their daily independence. 
As shown in the \Cref{table:home_life}, the overall process is divided into two parts:

\mypara{Housework} 
In complex home environments or during work and study scenarios, individuals with visual impairments often benefit from accurate recognition of object placement and attributes. 
This task evaluates the model’s ability to identify the positions and properties of everyday objects, including, but not limited to, their names, sizes, and colors, particularly in situations where multiple items are present and spatial relationships must be discerned.

\mypara{Leisure and Recreation} 
Beyond reading Braille books, many individuals with visual impairments also express a strong need to access standard printed books. 
To evaluate the model’s ability to understand physical printed books, we consider several key evaluation points, including correctly identifying the book title, navigating through the pages, recognizing the content on each page, and utilizing the table of contents to efficiently locate and access specific sections.

%----------------------------------
\subsubsection{Social Life Tasks}
%----------------------------------

This section simulates mobility tasks for visually impaired individuals during social activities, testing the model’s ability to support path recognition, transportation assistance, and destination guidance in real-world scenarios. As illustrated in the \Cref{table:social_life}
, the evaluation consists of three parts:

\mypara{Road Walking} 
Road walking evaluates the model’s capability to assist individuals with visual impairments during road walking tasks with environmental awareness and tactile path recognition. 
The evaluation includes describing the surrounding environment, providing navigation guidance along tactile paths, instructing the user on how to effectively use the cane, and offering real-time feedback. 
Such assistance can improve both the efficiency and safety of independent navigation for individuals with visual impairments.

\mypara{Transport} 
Traveling by vehicle poses inherent challenges for individuals with visual impairments. 
The ability to anticipate vehicle status and assess the surrounding environment in advance can significantly enhance travel safety and autonomy. 
This task evaluates the model’s capability to assist users in recognizing and approaching vehicles, covering aspects such as vehicle type, color, and motion state and supporting the full process of walking to the vehicle, opening the door, and boarding successfully.

\mypara{Reaching Destination} 
Once an individual with visual impairment has clearly identified the intended destination, determining the appropriate actions to take and the direction to move becomes a critical consideration. 
This task evaluates whether the model can provide comprehensive environmental guidance, accurately locate and assess the destination, and ultimately support the user in executing the required actions.

%----------------------------------
\subsection{\trainingdata}
\label{sec:safevid}
%----------------------------------

\begin{figure}[t]
    \centering
    \includegraphics[width=\columnwidth]{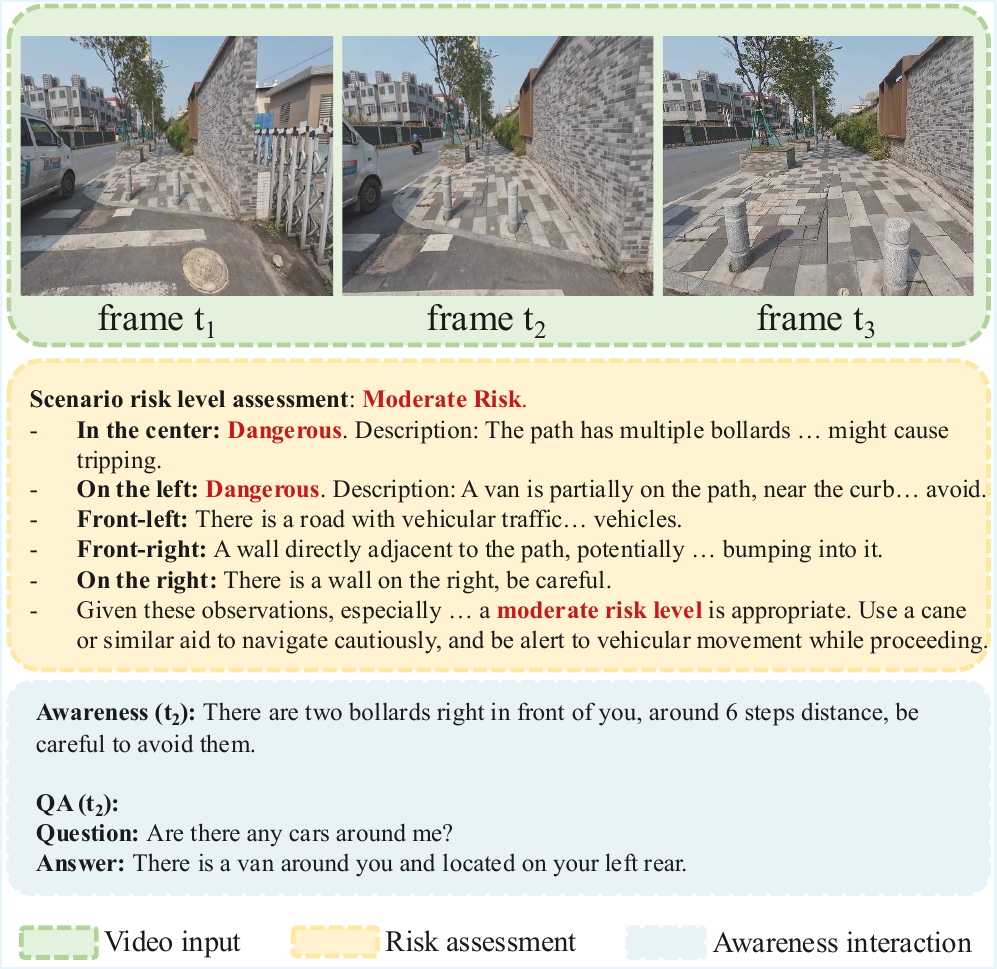}
    \caption{Example of SafeVid dataset.}
    \label{fig:safevid}
\end{figure}

To enable proactive risk alert functionality for VideoLLMs, we collect 2,205 high-definition 1080p video clips from real-world environments and annotate them based on the severity of potential environmental hazards present in each scene, as shown in~\Cref{fig:safevid}.

The collected videos cover typical locations visited by visually impaired individuals, such as bus stops, sidewalks, markets, and restaurants (See \Cref{fig:dataset} for more details).
Based on these data, we establish a risk-level classification system that categorizes each video according to the potential severity of future hazards. The system includes the following four levels:

\begin{itemize}[leftmargin=0pt]
    \item \textbf{No Risk}: There are no obstacles or potential dangers in the environment; the area is completely safe for passage.
    \item \textbf{Low Risk}: There are small soft/fixed obstacles, which may cause a slight tripping risk but not affect passage. It is recommended to use walking aids like canes for path detection.
    \item \textbf{Moderate Risk}: There are medium-sized hard obstacles whose sharp edges or fixed structures may cause physical injuries. Immediate avoidance is necessary.
    \item \textbf{High Risk}: There are large dangerous obstacles or sharp objects, mobile obstacles, and highly reflective/low contrast obstacles. Immediate evacuation and assistance are required.
\end{itemize}

\begin{table*}[ht]
  \centering
  \small
  \caption{
Model performance on the \bench benchmark across assistive tasks in three categories and two languages (EN: English and CH: Chinese).}
  \label{tab:accuracy}
  \begin{tabular}{ll*{3}{cc}}
    \toprule
    \multirow{2}{*}{\textbf{Category}} 
      & \multirow{2}{*}{\textbf{Task}}
      & \multicolumn{2}{c}{\textbf{GPT-4o}} 
      & \multicolumn{2}{c}{\textbf{Zhipu}} 
      & \multicolumn{2}{c}{\textbf{VITA‑1.5}} \\
    \cmidrule(lr){3-4}\cmidrule(lr){5-6}\cmidrule(lr){7-8}
      & 
      & \textbf{EN} & \textbf{CH} 
      & \textbf{EN} & \textbf{CH} 
      & \textbf{EN} & \textbf{CH} \\
    \midrule
    \multirow{4}{*}{Basic Skills}
      & Orientation Skills
        & 88.33\%   & 89.00\%
        & 66.67\%  & 100.00\%
        & 39.00\%  & 33.33\% \\
      & Guided Walking
        & 100.00\%    & 91.67\%
        & 55.67\%  & 74.00\%
        & 16.67\%    & 27.33\% \\
      & Independent Walking
        & 100.00\%  & 100.00\%
        & 61.00\%    & 50.00\%
        & 55.67\%   & 58.33\% \\
      & Cane Techniques
        & 100.00\%     & 100.00\%
        & 16.67\%  & 83.33\%
        & 66.67\%  & 100.00\% \\
    \midrule
    \multirow{2}{*}{Home Life Tasks}
      & Housework
        & 100.00\%    & 100.00\%
        & 100.00\%    & 66.67\%
        & 50.00\%    & 100.00\% \\
      & Leisure and Recreation
        & 95.33\%    & 100.00\%
        & 55.33\%     & 70.67\%
        & 68.33\%     & 65.00\% \\
    \midrule
    \multirow{3}{*}{Social Life Tasks}
      & Road Walking
        & 100.00\%  & 100.00\%
        & 33.33\%  & 83.33\%
        & 66.67\%   & 83.33\% \\
      & Transport
        & 50.00\%  & 66.67\%
        & 41.67\%     & 33.33\%
        & 33.33\% & 50.00\% \\
      & Reaching Destination
        & 100.00\%    & 100.00\%
        & 66.67\%     & 100.00\%
        & 83.33\%   & 33.33\% \\
    \midrule
    \textbf{Average} 
      & -
      & 92.63\% & 94.15\%
      & 55.22\% & 73.48\%
      & 53.30\% & 61.18\% \\
    \bottomrule
  \end{tabular}
\end{table*}

%----------------------------------
\section{Experimental Setups}
%----------------------------------

\mypara{VideoLLMs Evaluation} 
We evaluate the assistive potential of three SOTA VideoLLMs, comprising both production-level and open-source systems, in supporting severely visually impaired individuals:
(1) GPT-4o executes through the ChatGPT iOS application (v1.2025.043); (2) Zhipu is used through its iOS client (v2.7.0); (3) VITA-1.5 can be found in its Github.~\footnote{\url{https://github.com/VITA-MLLM/VITA}.}

\mypara{Input Devices}
In this paper, we use an ``iPhone 13 Pro Max in iOS 18.3.1 system''~\footnote{ \url{https://www.apple.com/by/iphone-13-pro/specs/}.} as the primary testing device.
The app/website with the VideoLLMs are executed on this device, and first-person video data is captured using its built-in screen recording function. 
To obtain a more experimental record, we adopt a dual-device recording setup by introducing a ``DJI Action 4''~\footnote{\url{https://www.dji.com/cn/osmo-action-4}.} camera mounted on a stabilizing rig to simultaneously capture video of a third-person perspective.

The data collection for \trainingdata is conducted entirely using the ``DJI Action 4'' camera, which provides stable and high-resolution environmental video footage.

\mypara{Evaluation Metrics}
To comprehensively evaluate the performance of VideoLLMs in real-world tasks, we assess each task $i$ based on four key metrics:

Task Success Rate (TSR) measures the effectiveness of the model's responses in helping users accomplish the intended task. 
The task success rate over $N$ tasks is defined as:
\begin{equation}
\text{TSR} = \frac{1}{N} \sum_{i=1}^{N} s_i,
\end{equation}
where $s_i \in \{0, 1\}$ denote whether task $i$ is successful.

Average Prompt Cost (APC) quantifies the number of user inputs required for the model to complete a task, reflecting the efficiency of interaction. 
The average prompt cost is computed as:
\begin{equation}
\text{APC} = \frac{1}{N} \sum_{i=1}^{N} P_i,
\end{equation}
Where the $P_i$ represent the number of prompts used in task $i$.

Average Response Latency (ARL) evaluates the model's responsiveness, i.e., the average time delay between a user prompt and the corresponding model response. 
The ARL is defined as:
\begin{equation}
\text{ARL} = \frac{1}{N} \sum_{i=1}^{N} \left( t_i^{\text{resp}} - t_i^{\text{prompt}} \right),
\end{equation}
where $t_i^{\text{prompt}}$ be the time when the prompt is issued, and $t_i^{\text{resp}}$ the time when the response is received. 

Language Consistency (LC) measures whether the model's output remains in the same language as the user’s input. 
The LC is defined as:
\begin{equation}
\text{LC} = \frac{1}{N} \sum_{i=1}^{N}\mathbbm{1} \left( L_i^{\text{resp}} = L_i^{\text{prompt}} \right),
\end{equation}
where $\mathbbm{1}(\cdot)$ is the indicator function that returns 1 if the condition holds, and 0 otherwise. $L_i^{\text{prompt}}$ and $L_i^{\text{resp}}$ are languages of the input prompt and the model response, respectively.

\mypara{Training Details in \trainingdata}
To enable the model to alert users to potential dangers proactively, we fine-tune the VITA-1.5 model with \trainingdata, which has 1,204 video clips.
The model is fine-tuned for 2 epochs with a batch size of 4, a learning rate of 1e-5, and the cosine learning rate scheduler. 
The training is performed on six Nvidia L20 GPUs.

\begin{table*}[t]
    \centering
    \caption{
        Participant demographics, including age, visual status, task success rates in closed-world (TSR (C)) and open-world (TSR (O)) scenarios, and satisfaction scores. 
        Visual status abbreviations: B = Blind, SVI = Severely Visually Impaired, S = Sighted; onset abbreviations: Youth = Since Youth, Later = Later In Life, BF = Blindfolded.
    }
    \label{tab:participant_data}
    \setlength{\tabcolsep}{2.5pt}
    \renewcommand{\arraystretch}{1.1}
    \scriptsize
    \begin{tabularx}{\textwidth}{l c c l l l l >{\raggedright\arraybackslash}X c c l}
        \toprule
        \textbf{ID} & \textbf{Age} & \textbf{Gender} & \textbf{Visual Status} & \textbf{Onset} & \textbf{Education} & \textbf{Occupation} & \textbf{AI Usage} & \textbf{TSR (C)} & \textbf{TSR (O)} & \textbf{Satisfaction Scores} \\
        \midrule
        U1 & 90 & Male & B   & Youth & Primary school              & Retired / home living      & No prior; needs guided onboarding                    & 60.00\%  & 75.00\%  & 7 (Generally satisfied)   \\
        \midrule
        U2 & 35 & Male & B   & Youth & Vocational secondary        & Manual work                & Screen reader only; willing to adopt                 & 100.00\% & 66.67\% & 8 (Generally satisfied)   \\
        \midrule
        U3 & 30 & Female & SVI & Youth & Vocational secondary       & Massage therapist          & Limited; object/scene for tools/traffic             & 75.00\%  & 50.00\% & 9 (Highly satisfied)      \\
        \midrule
        U4 & 21 & Female & SVI & Youth & University (undergrad)     & Student                    & Occasional; study assistance                        & 80.00\%  & 83.33\% & 7 (Generally satisfied)   \\
        \midrule
        U5 & 27 & Female & B   & Later & Junior college (associate) & Family-assisted living     & Yes; object/scene, family-assisted                  & 80.00\%  & 100.00\% & 6 (Generally satisfied)   \\
        \midrule
        U6 & 24 & Female & S   & BF    & University                  & Student                    & Regular chat with study assistance                  & 100.00\% & 100.00\% & 7 (Generally satisfied)   \\
        \midrule
        U7 & 22 & Female & S   & BF    & University                  & Student                    & Regular chat; daily queries                         & 100.00\% & 100.00\% & 8 (Generally satisfied)   \\
        \midrule
        U8 & 24 & Female & S   & BF    & University                  & Student                    & Regular; reading augmentation                       & 100.00\% & 80.00\% & 8 (Generally satisfied)   \\
        \bottomrule
    \end{tabularx}
\end{table*}

%----------------------------------
\section{Evaluation}
%----------------------------------

%----------------------------------
\subsection{Performance Evaluation}
%----------------------------------

Based on our constructed benchmark dataset \bench, we evaluate three models: GPT-4o, Zhipu, and VITA-1.5. 
The results are averaged from tests conducted by three individuals wearing blindfolds to simulate visual impairments (as shown~\Cref{fig:dataset}).

\Cref{tab:accuracy} presents the TSR across the three models.
GPT-4o performs the best in most tasks, especially within the ``Home Life Tasks'' and ``Basic Skills'', where it achieves 100\% TSR on multiple tasks. 
Further, the gap between its TSR in English and Chinese is minimal, indicating balanced multilingual capabilities.
Zhipu performs slightly below GPT-4o but still shows strong results in both task sets, achieving an average TSR of 55.22\% and 73.48\%, respectively. 
However, its performance varies significantly across tasks. For example, it records a low success rate of only 33.33\% in ``Road Walking (EN)'' and just 16.67\% in ``Cane Techniques (EN)'', indicating challenges in handling certain task-specific scenarios.
VITA-1.5 shows the weakest overall performance, with a TSR of only 53.30\% in English and 61.18\% in Chinese. 
The model demonstrates inconsistent behavior across languages for different tasks. 
For instance, it achieves 100\% TSR in ``Housework (CH)'', outperforming its 65.00\% in ``Housework (EN)'', but only 33.33\% in ``Reaching Destination (EN)'', reflecting unstable multilingual task performance.

\Cref{tab:model_metrics} shows the models' performance across various evaluation metrics.
Overall, GPT-4o ranks the first, achieving the highest TSR and LC, with the lowest ARL in English. It also records the lowest APC in CH tasks.
\begin{table}{r}
\centering
\caption{Comparison of VideoLLMs performance across four evaluation metrics.}
\label{tab:model_metrics}
\resizebox{\linewidth}{!}{
\begin{tabular}{lcccccc} 
\toprule
\textbf{Metric}
  & \multicolumn{2}{c}{\textbf{GPT-4o}}
  & \multicolumn{2}{c}{\textbf{Zhipu}}
  & \multicolumn{2}{c}{\textbf{VITA-1.5}} \\
\cmidrule(lr){2-3} \cmidrule(lr){4-5} \cmidrule(lr){6-7}
  & \textbf{EN} & \textbf{CH}
  & \textbf{EN} & \textbf{CH}
  & \textbf{EN} & \textbf{CH} \\
\midrule
\textbf{TSR$\uparrow$}
  & 92.63\% & 94.15\%
  & 55.22\% & 73.48\%
  & 53.30\% & 61.18\% \\
\textbf{ARL$\downarrow$}
  & 2.36 & 2.28
  & 2.66 & 2.36
  & 2.37 & 2.26 \\
\textbf{APC$\downarrow$}
  & 3.00 & 2.85
  & 3.52 & 3.52
  & 2.78 & 3.48 \\
\textbf{LC$\uparrow$}
  & 100.00\% & 100.00\%
  & 76.81\%  & 100.00\%
  & 93.33\%  & 100.00\% \\
\bottomrule
\end{tabular}%
}
\end{table}
Zhipu ranks the second, with a solid average TSR of 73.48\% under CH input and moderate ARL (2.66) and APC (3.52). However, it struggles to maintain consistent language output, often responding in Chinese to EN prompts.
VITA-1.5 performs the worst overall across all models.

\mypara{Error Case Analysis}
The following section presents two error cases encountered during the evaluation process, with a detailed analysis of their underlying causes.

In error case 1, GPT-4o demonstrates spatial reasoning limitations, mislocating a cane positioned on the right side of the room by describing it as being on the left and near a distant black chair. 
This reflects confusion between object co-occurrence and actual spatial proximity, as well as unreliable left-right orientation.
In error case 2, Zhipu incorrectly localizes the user at the top of a staircase despite clear contextual cues indicating the bottom position.
The model further described a flat hallway ahead, omitting the presence of stairs. 
This reflects a failure in spatial grounding and may result in safety risks in assistive navigation contexts.

\begin{figure*}[t]
    \centering
    \includegraphics[width=1\textwidth]{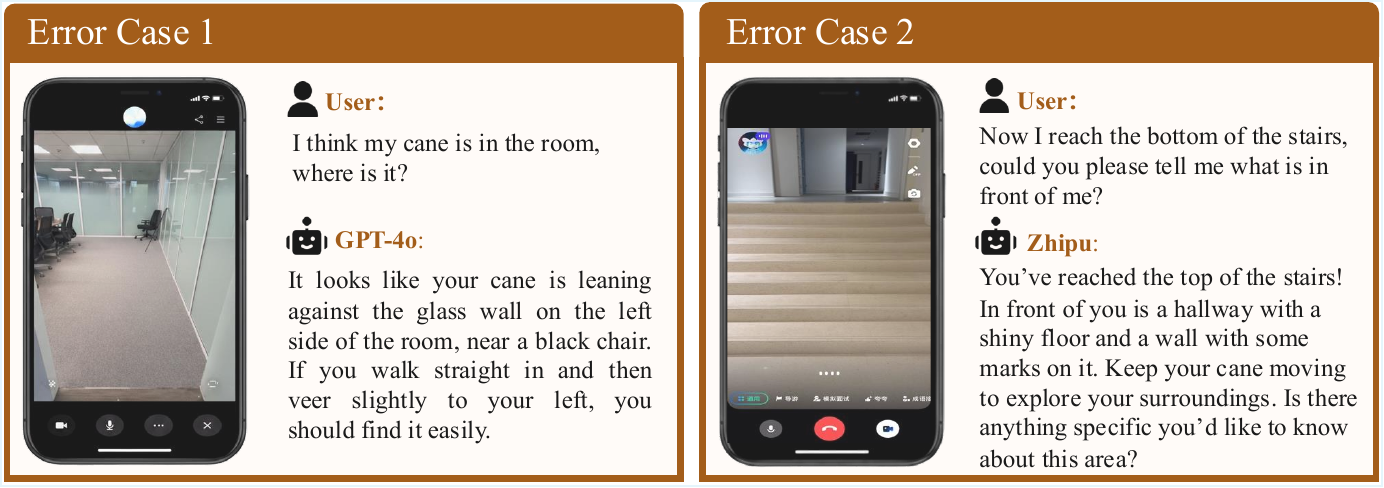}
    \caption{Example of error cases.}
    \label{fig:error_case}
\end{figure*}

\begin{table}[t]
    \small
    \caption{Satisfaction scale and performance interpretation.}
    \label{tab:satisfaction}
    \centering
    \begin{tabularx}{\columnwidth}{c|X}
        \hline
        \textbf{Score (1--10)} & \textbf{Description} \\
        \hline
        1--3  & Significantly dissatisfied; unable to complete assigned tasks effectively. \\
        4--5  & Marginally satisfied; able to complete only basic tasks with limited proficiency. \\
        6--8  & Generally satisfied; capable of handling moderately complex tasks with reasonable competence. \\
        9--10 & Highly satisfied; fully proficient in completing all assigned tasks efficiently. \\
        \hline
    \end{tabularx}
\end{table}

%----------------------------------
\subsection{User Study}
%----------------------------------

To further gather subjective evaluations from users with different backgrounds and situations on VideoLLM-based assistance, we conduct a user study in which participants are invited to complete a series of real-world tasks. During the process, we record their feedback and performance to better understand their experiences with the system.

%----------------------------------
\subsubsection{Setup}
%----------------------------------

In the setup of the user study, we consider 4 perspectives, including participants, input devices and model, task, and performance metrics.

\mypara{Participants}
We recruit 8 volunteers through social media platforms, including RedNotes and Weibo, to participate in the user study in either an online or in-person manner. 
Each participant is compensated at an hourly rate of 15 dollars. 
As shown in~\Cref{tab:participant_data}, the participants are in a diverse range of ages (from 21 to 90). 
Among them, U1 to U5 are individuals who are severely visually impaired and are almost completely blind, while U6 to U8 are sighted individuals who wear blindfolds during study to simulate visual impairment.

\mypara{Input Devices and Model}
The experiments are conducted on participants’ personal smartphones using the same version of Zhipu.
Zhipu is selected because it is the most easily accessible and usable model for the general public, as GPT-4o requires a paid subscription and VITA-1.5 needs deployment.

\mypara{Task}
The evaluation tasks are divided into two categories: closed-world tasks and open-world tasks (user-defined goals).
Closed-world tasks include: (1) Finding a specific item in a room;
(2) Identifying a book and helping to determine its title and table of contents;
(3) Going upstairs, with the model guiding users by describing step height and assisting with navigation;
(4) Entering a room and having the model describe the surroundings;
(5) Walking outdoors, with the model providing feedback on road conditions.
Open-world tasks include finding multiple similar objects, crossing the street, finding medicine bottles, judging road conditions and crossing traffic lights, finding tools in a workspace, assisting with learning, experiencing recreational activities, and expanding the scope of learning.

\mypara{Evaluation Metrics}
We use only TSR as the main metric, since some participants are not comfortable with recording video, making it difficult to collect other measurements.
In addition, we use a Satisfaction score to capture overall user experience, which is rated by each participant after completing the tasks.
The scoring criteria are detailed in~\Cref{tab:satisfaction}.

%----------------------------------
\subsubsection{Evaluation}
%----------------------------------

We present the results of the user study along with participants' feedback and insights based on their interaction experiences with the system.

\mypara{Performance Analysis}
As shown in~\Cref{tab:participant_data}, we find that participants achieve high TSR in both closed-world and open-world settings. 
Notably, users U2, U6, U7, and U8 achieve a closed-world TSR of 100.00\%.
Besides, users U5, U4, and U3 achieve closed-world TSRs of 80.00\%, 80.00\%, and 75.00\%, respectively.
In the open-world settings, users U5, U6, and U7 achieve a TSR of 100.00\%.
Also, users U4, U8, and U1 attain TSRs of 83.00\%, 80.00\%, and 75.00\%, respectively. 
These results suggest that Zhipu performs well in most scenarios, demonstrating high TSR across a broad range of tasks.
However, a notable exception is observed with user U1 and U3, who have  60.00\% and 50\% closed-world TSR.
This lower performance is likely due to Zhipu’s limitations in distinguishing between similar objects.
Despite these exceptions, all users acknowledge the capabilities of Zhipu and are generally satisfied with the model, as demonstrated by their satisfaction scores are all above $6$. This indicates that, overall, users find the model capable of handling complex tasks to a reasonable competence.

%----------------------------------
\subsubsection{Feedback from Participants}
%----------------------------------

We organize participants' feedback into four themes shown below.

\mypara{Everyday Use Cases: Self-care and  Environmental Adaptation}
Participants’ assistive needs concentrate on activities of daily living, including meal preparation (U1), medication management (U1), personal hygiene (U1), and mobility (U2, U3, U5). Beyond self-care, participants also seek support for recreational activities (U5) and educational tasks (U4, U7). Across these contexts, two capabilities consistently surface as high-priority: object localization (U1--U6) and broader environmental awareness (U3, U4, U6, U8).

\mypara{Wayfinding and Spatial Understanding}
When seeking technological assistance for navigation, participants emphasize tasks such as finding directions (U1), recognizing rooms and locations (U3, U5), crossing streets (U1, U2), and interpreting traffic signals (U2, U3). These reports underscore object localization as a prerequisite for safe action in both indoor and outdoor spaces, and highlight the need for detailed, situation-aware scene descriptions beyond coarse labels.

\mypara{Failure Modes and Interaction  Pain Points}
Field tests reveal inconsistent performance across scenarios. Errors are most prominent in crowded or visually complex settings, with inaccurate object or environment descriptions (U4, U6, U8). Participants also encounter imprecise spatial language (U4, U6), incorrect or incomplete object recognition (U4, U6, U8), insufficient descriptive detail for decision making (U4, U8), and difficulty translating visual information into actionable steps (U4, U6). Additional issues include occasional incorrect confirmation responses (U4), and challenges with voice interaction in public—privacy concerns and unstable automatic speech recognition (ASR) in noisy environments (U4, U8). Beyond perception and interaction, users mention slow interaction speed (U1, U8), difficulty following long or complex instructions (U7, U8), and the need for persistent reminders or guidance (U8). Figure~\cref{fig:common-problems} summarizes the prevalence of these issues across participants.

\mypara{Deep-dive Interview of U4}
In a longer follow-up interview, U4 emphasizes two concerns frequently encountered in outdoor use. First, speech-based interaction is often impractical due to privacy exposure and unreliable recognition under noise; U4 suggests a Frequency-Answer-Question (FAQ-like module for quick access to situation-specific information without extensive voice input. Second, and the most critical concern is that recognition alone (e.g., detecting a staircase) rarely suffices; users need follow-up, step-by-step guidance to complete tasks safely.

\mypara{Design Implications}
Taken together, these findings point to several design opportunities: (1) privacy-aware, low-audibility interaction modes (earcons/haptics) and robust ASR for noisy environments; (2) fast access to context-sensitive “FAQ” prompts or templates; (3) action-oriented outputs that translate recognition into procedural steps with safety checks; (4) clearer and more calibrated spatial descriptions (with consistent frames of reference and distance cues); (5) progressive disclosure and user-controlled pacing/summarization for long instructions; and (6) persistent guidance and reminders for safety-critical tasks.

\begin{figure}[t]
  \centering
  \includegraphics[width=0.9\columnwidth]{Image/failure_points.png}
  \caption{Summary of common failure modes and interaction pain points reported by participants.}
  \label{fig:common-problems}
\end{figure}

%----------------------------------
\subsection{Challenges and Future Directions}
%----------------------------------

Based on user feedback analysis and benchmark evaluations, we have identified several key challenges currently faced by VideoLLMs, which are outlined as follows:

\mypara{Challenge 1: Limited Performance of VideoLLMs}
Although the current VideoLLMs have shown strong capabilities in real-time video understanding, user feedback and benchmark evaluations have revealed several key issues that need to be addressed:
1. \textbf{Interaction quality defects in noisy environments}
In complex auditory scenarios, the reliability of the voice interaction of VideoLLMs is significantly reduced.
Specifically, when the user is in a noisy environment, the model has difficulty in effectively extracting the user's audio input.
User feedback shows that noise interference mainly comes from two aspects: environmental background sound (such as traffic noise and crowd conversation) and clothing friction caused by user body movements (especially jackets close to mobile phones).
2. \textbf{Insufficient adaptability to low-light environments:}
The current VideoLLMs are significantly limited in dim light or night scenes.
Its visual recognition ability will be perceptibly reduced. For example, user feedback pointed out that in night street walking scenes, models such as Zhipu have difficulty accurately identifying surrounding objects and location information.
3. \textbf{Fine-grained object recognition defects:}
VideoLLMs lack the ability to distinguish objects with similar visual features.
For example, the misjudgment rate of different cup types is high, and it is difficult to capture and describe subtle texture differences (based on our initial trial).
4. \textbf{Object description credibility issues:}
User study feedback that the VideoLLMs have a ``hallucination'' phenomenon.
For example, Zhipu describes objects that do not exist in the environment in the test case.
5. \textbf{Lack of consistency in language interaction:}
The VideoLLMs have the problem of out-of-control language switching. 
For example, when users query in English, Zhipu sometimes will respond in Chinese.
The above problems expose the defects of the VideoLLMs in three dimensions: environmental adaptability, visual cognition accuracy, and interaction certainty.

In general, although VideoLLMs have shown their ability in handling these tasks, more efforts have to be made to achieve reliable deployment in real-world scenarios.

\mypara{Challenge 2: Lack of Proactive Perception Capability}
Current VideoLLMs generally adopt a passive response mechanism; that is, the model response is triggered only when the user initiates a query.
That means the model cannot proactively anticipate dynamic environmental risks (such as sudden obstacles), resulting in its inability to issue warnings to users before danger occurs.
To overcome this limitation, we propose \trainingdata to integrate proactive environmental perception capabilities into model training. For the details of our method, please refer to the~\Cref{sec:reminder}.

\mypara{Challenge 3: Limited Usability in Practice}
In our experiments, interactions with VideoLLMs are conducted using smartphones that are worn around the neck and run the official apps. However, since the existing VideoLLMs are not optimized for the special needs of visually impaired users, the following difficulties are exposed in actual use:
1. \textbf{Hardware interaction design flaws:}
The user reports that the significant sense of weight caused by the long-term wearing of the device (smartphone) affects the willingness to continue using it. 
To address this, a potential improvement could be to explore VideoLLMs integration solutions with lightweight wearable devices such as smart glasses;
2. \textbf{Lack of privacy protection mechanism:}
Participants have psychological pressure to be eavesdropped when speaking personal information in public places. To address this, future research could focus on developing non-voice vibration/tactile feedback interaction channels to protect user privacy;
3. \textbf{Lack intelligent customer service:}
Users suggest that VideoLLMs can adopt interaction patterns from intelligent customer service systems. 
For example, offering proactive suggestions or providing answers to commonly ask questions can streamline user interactions, enabling them to access information more efficiently and significantly improving both the speed of interaction and the overall user experience.

\mypara{Challenge 4: Security Issues of VideoLLMs}
Previous studies show that LLMs and VLMs generally have security vulnerabilities and are susceptible to backdoor attacks~\cite{zhang2024instruction,DBLP:journals/corr/abs-2411-17453} and jailbreak attacks~\cite{sun2025survive,DBLP:journals/corr/abs-2407-04295,DBLP:journals/corr/abs-2311-05608,DBLP:journals/corr/abs-2502-21059,DBLP:journals/corr/abs-2411-19530}. 
These vulnerabilities may pose potential risks to visually impaired individuals using such systems.
Although security is not the primary focus of this study, it remains a critical concern. 
To explore this issue, we conduct a preliminary security evaluation by performing a jailbreak attack on the Zhipu model using the FC-Attack~\cite{DBLP:journals/corr/abs-2502-21059}. 
We find that the model could be successfully jailbroken\footnote{The vulnerability has been promptly reported to the service provider for remediation.}.
Therefore, we argue that before VideoLLMs are widely deployed in assistive applications for visually impaired users, their security risks must be thoroughly studied and effectively addressed.

%----------------------------------
\subsection{Proactive Reminder Ability}
\label{sec:reminder}
%----------------------------------

As described in~\Cref{sec:safevid}, we construct a video-text paired dataset, \trainingdata, to capture potential risk levels in various environments.
Based on this dataset, we modify the VITA-1.5 framework to enable the model to proactively analyze video content through timed polling and periodically generate descriptions of complicated environmental hazards.
The model's performance is evaluated using Accuracy, BLEU\cite{DBLP:conf/acl/PapineniRWZ02}, and ROUGE-L\cite{lin-2004-rouge}.
We define accuracy as the fraction of examples for which the model's predictions, obtained before and after fine-tuning, coincide with the ground truth labels.

As shown in ~\Cref{tab:vita_results}, the original VITA-1.5 achieves an accuracy of only 25.00\% on the environmental risk recognition task. After task-specific fine-tuning, the model's accuracy improves to 76.00\%, a nearly threefold improvement.
Also, the fine-tuned model yields a BLEU score of 31.41 and a ROUGE-L score of 40.06, where higher values on both metrics indicate better performance.
From ~\cref{fig:Confusion Matrix of Risk Levels}, the fine-tuned model is noticeably better at recognizing the primary risk: accuracy improves for No Risk (from 44\% to 57\%), Low Risk (from 20\% to 68\%), and Moderate Risk (from 29\% to 100\%). 
The confusion matrix indicates that the previous conservative tendency to downgrade samples has largely been corrected; remaining errors are mostly adjacent confusions between Low Risk and Moderate Risk. 
As a result, the model more reliably separates low from medium risk, reducing safety risks from missed or incorrect classifications.
This implies that fine-tuning with high-quality dataset could enhance the VideoLLM's proactive reminder ability significantly, which could serve as a valuable future research direction.

\begin{table}[htbp]
\centering
\small
\caption{Performance of vanilla and fine-tuned VITA-1.5.}
\label{tab:vita_results}
\begin{tabular}{lccc}
\toprule
\textbf{Model} & \textbf{Accuracy} & \textbf{BLEU} & \textbf{ROUGE-L} \\
\midrule
VITA-1.5-Vanilla & 25.00\% &2.76 &18.35 \\
VITA-1.5-Fine-Tuned & 76.00\% & 31.41 & 40.06 \\
\bottomrule
\end{tabular}
\end{table}

%----------------------------------
\section{Limitations}
%----------------------------------

We conduct a comprehensive evaluation of three SOTA VideoLLMs for assisting individuals with visual impairments through both systematic benchmarking and a user study.
It still has some limitations:

\begin{itemize}
    \item \textbf{Limited model and language coverage:} 
    To the best of our knowledge, our study evaluates all the real-time VideoLLMs but only focuses on two languages. 
    Future directions include incorporating new models and diverse languages to better support global accessibility.
    \item \textbf{Restricted geographic and participant diversity:} Due to resource constraints, the proposed benchmark (\bench) currently covers only a limited set of environments and countries. Similarly, the number of participants in the user study was relatively small.
    Expanding both the geographic scope and the participants from diverse cultural and linguistic backgrounds will be crucial for building a more representative and generalizable evaluation.
    \item \textbf{Unresolved challenges:} While our study highlights key challenges encountered in the real-world use of VideoLLMs for assistive tasks, many of these challenges remain unsolved. Addressing them constitutes important directions for future research.
\end{itemize}

%----------------------------------
\section{Conclusion}
%----------------------------------

This work is the first to focus on leveraging real-time interactive VideoLLMs for daily assistance to individuals with visual impairments.
Building on previous research and insights from visually impaired volunteers, we construct an evaluation dataset named \bench, and systematically evaluate three SOTA real-time VideoLLMs. 
Experimental results show that GPT-4o achieves the highest task success rate, while VITA-1.5 performs the worst.
We then recruit eight volunteers to conduct real-world testing using the Zhipu VideoLLMs. 
Overall, participants express a high level of satisfaction with the system while also providing valuable feedback and highlighting several challenges. 
Based on these findings, we summarize key challenges currently faced by VideoLLMs in assistive applications for the visually impaired.
To equip VideoLLMs with proactive risk awareness, we build a dedicated dataset (\trainingdata) and fine-tune VITA-1.5, boosting risk detection accuracy to 76.00\% and demonstrating its assistive potential.

\newpage
\bibliographystyle{plain}
\bibliography{sample-base}

\appendix
\begin{figure}[htbp]
  \centering
  \begin{subfigure}{0.48\textwidth}
    \centering
    \includegraphics[width=\linewidth]{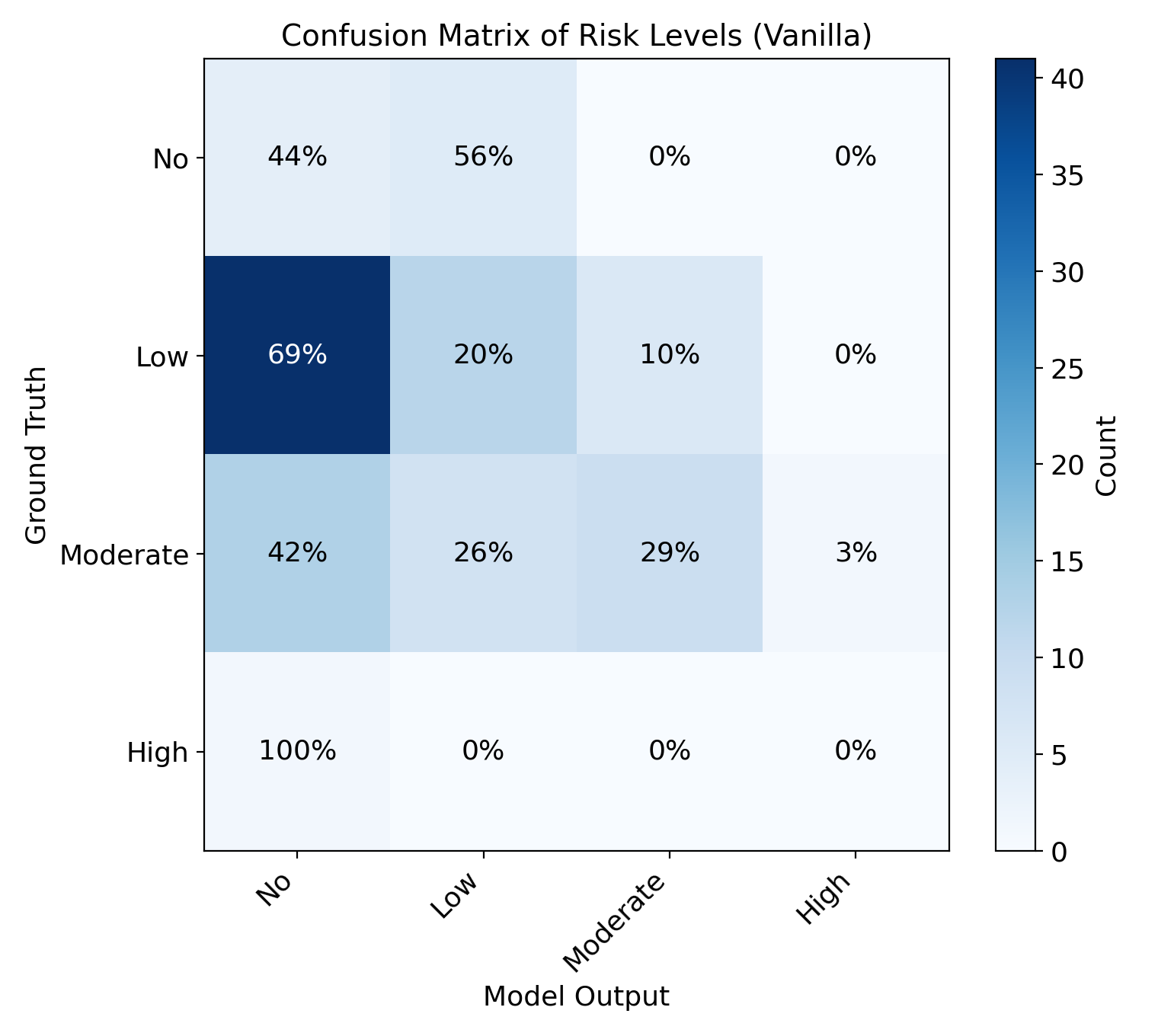}
    \caption{Vanilla}
    \label{fig:Confusion Matrix of Risk Levels Vanilla}
  \end{subfigure}\hfill
  \begin{subfigure}{0.48\textwidth}
    \centering
    \includegraphics[width=\linewidth]{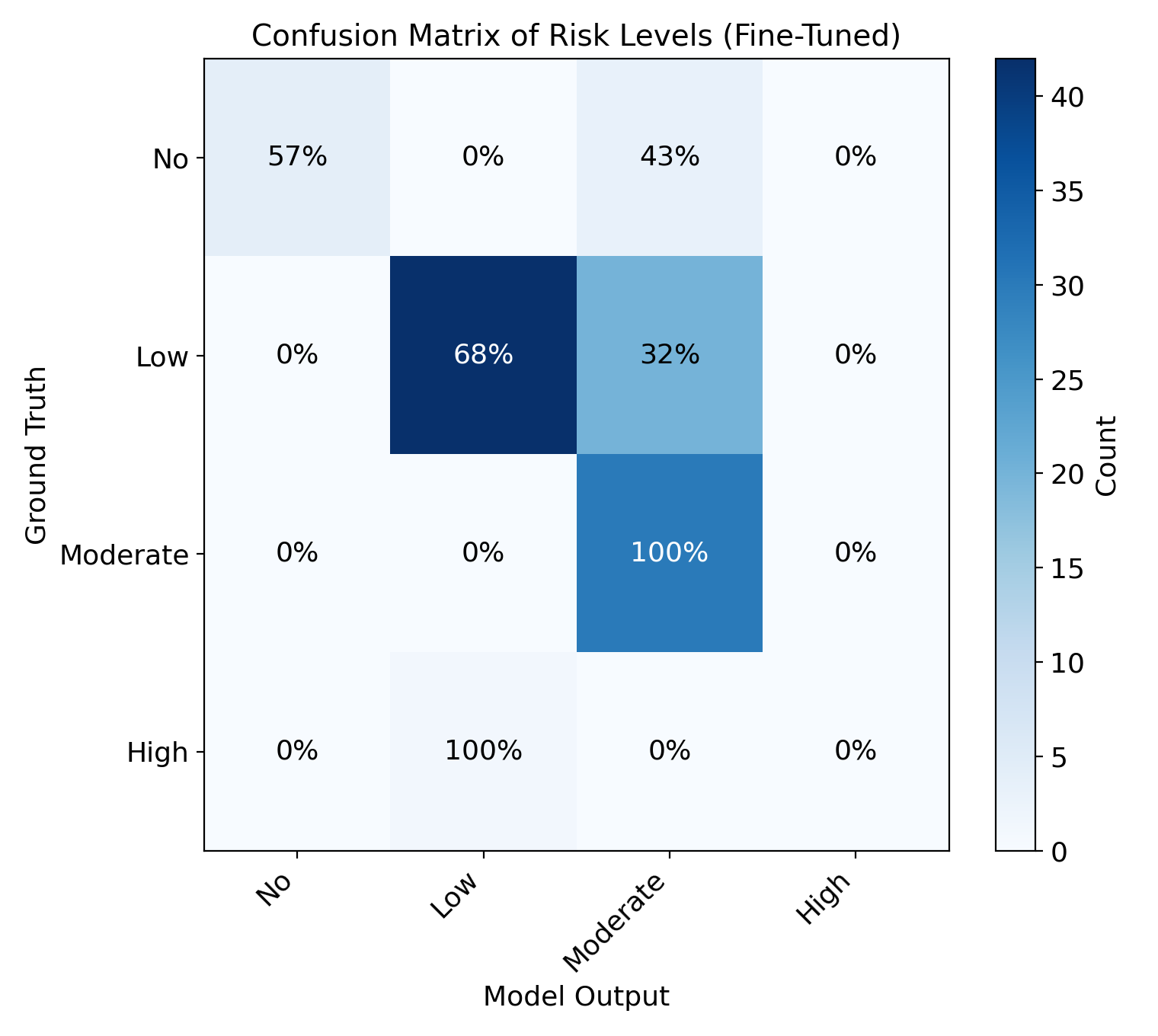}
    \caption{Fine-Tuned}
    \label{fig:Confusion Matrix of Risk Levels Fine Tuned}
  \end{subfigure}
  \caption{Confusion matrix of vanilla and fine-tuned VITA-1.5.}
  \label{fig:Confusion Matrix of Risk Levels}
\end{figure}

\end{document}